\documentclass[journal,transmag]{IEEEtran}
\ifCLASSINFOpdf
\else
\fi
\usepackage{url}

\usepackage{times}
\usepackage{epsfig}
\usepackage{graphicx}
\usepackage{amsmath}
\usepackage{amssymb}
\usepackage{subfigure}
\usepackage{upgreek}
\usepackage{multirow}
\usepackage{color}
\usepackage{bm}
\DeclareMathOperator*{\argmin}{arg\,min}
\usepackage{arydshln}
\usepackage{cite}
\usepackage{pdfpages}
\usepackage{blindtext}

\let\emptyset\varnothing

\usepackage{amsthm,xpatch}

\xpatchcmd{\proof}{\hskip\labelsep}{\hskip3\labelsep}{}{} 
\newtheorem{theorem}{Theorem}

\usepackage{caption}
\usepackage{xcolor}

\usepackage[pagebackref=true,breaklinks=true,letterpaper=true,colorlinks,bookmarks=false]{hyperref}

\begin{document}
%

\title{External Prior Guided Internal Prior Learning 
\\
for Real-World Noisy Image Denoising}


\author{
\IEEEauthorblockN{Jun Xu$^{1}$,~\IEEEmembership{Student Member,~IEEE},
Lei Zhang\textsuperscript{1,*}\thanks{*This research is supported by the HK RGC GRF grant (PolyU 152124/15E).},~\IEEEmembership{Fellow,~IEEE},
and
David Zhang$^{1,2}$,~\IEEEmembership{Fellow,~IEEE}
}
\IEEEauthorblockA{$^{1}$Department of Computing,
The Hong Kong Polytechnic University, Hong Kong SAR, China
\\
$^{2}$School of Science and Engineering, The Chinese University of Hong Kong (Shenzhen), Shenzhen, China
}


}

\markboth{}%
{Shell \MakeLowercase{\textit{et al.}}: Bare Demo of IEEEtran.cls for IEEE Transactions on Magnetics Journals}
%



\IEEEtitleabstractindextext{%
\begin{abstract}
Abstract: Most of existing image denoising methods learn image priors from either external data or the noisy image itself to remove noise. However, priors learned from external data may not be adaptive to the image to be denoised, while priors learned from the given noisy image may not be accurate due to the interference of corrupted noise. Meanwhile, the noise in real-world noisy images is very complex, which is hard to be described by simple distributions such as Gaussian distribution, making real-world noisy image denoising a very challenging problem. We propose to exploit the information in both external data and the given noisy image, and develop an external prior guided internal prior learning method for real-world noisy image denoising. We first learn external priors from an independent set of clean natural images. With the aid of learned external priors, we then learn internal priors from the given noisy image to refine the prior model. The external and internal priors are formulated as a set of orthogonal dictionaries to efficiently reconstruct the desired image. Extensive experiments are performed on several real-world noisy image datasets. The proposed method demonstrates highly competitive denoising performance, outperforming state-of-the-art denoising methods including those designed for real-world noisy images.
\end{abstract}

\begin{IEEEkeywords}
Image Denoising, Real-World Noisy Image, Image Prior Learning, Guided Dictionary Learning.
\end{IEEEkeywords}}

\maketitle

\IEEEdisplaynontitleabstractindextext

%
\IEEEpeerreviewmaketitle

\vspace{-1mm}
\section{Introduction}

\IEEEPARstart{I}{mage} denoising is a crucial and indispensable step to improve image quality in digital imaging systems. In particular, with the decrease of size of CMOS/CCD sensors, image is more easily to be corrupted by noise and hence denoising is becoming increasingly important for high resolution imaging. The problem of image denoising has been extensively studied in literature and numerous image denoising methods
\cite{bayesshrink,curvelet,ksvd,lssc,ncsr,bm3d,cbm3d,
zhou2012nonparametric,Tomasi1998,blsgsm,nlm,nlbayes,wnnm,pgpd,foe,epll,
combexin,external,tid,
mlp,xie2012image,dncnn,
barbu2009training,csf,chen2015learning,Fadili,salmon2014,
Foipractical,Luisier,Makitalo2013Optimal,Montagner,
jiang2014mixed,Hu2016,xuaccv2016,
fullyblind,rabie2005robust,Liu2008,almapg,noiseclinic,
ncwebsite,Zhu_2016_CVPR,crosschannel2016,mcwnnm,neatimage}
have been proposed in the past decades. Most of existing denoising methods focus on the scenario of additive white Gaussian noise (AWGN) 
\cite{bayesshrink,curvelet,ksvd,lssc,ncsr,bm3d,cbm3d,
zhou2012nonparametric,Tomasi1998,blsgsm,nlm,nlbayes,wnnm,pgpd,combexin,external,tid,foe,epll,
mlp,xie2012image,dncnn,barbu2009training,csf,chen2015learning}, where the observed noisy image $\mathbf{y}$ is modeled as the addition of clean image $\mathbf{x}$ and AWGN $\mathbf{n}$, i.e., $\mathbf{y}=\mathbf{x}+\mathbf{n}$. There are also methods proposed for removing Poisson noise \cite{Fadili,salmon2014}, mixed Poisson and Gaussian noise \cite{Foipractical,Luisier,Makitalo2013Optimal,Montagner}, mixed Gaussian and impulse noise \cite{jiang2014mixed,Hu2016,xuaccv2016}, and realistic noise in real photography \cite{fullyblind,rabie2005robust,Liu2008,almapg,Zhu_2016_CVPR,noiseclinic,
ncwebsite,crosschannel2016,neatimage,mcwnnm}.

Natural images have many properties, such as sparsity and nonlocal self-similarity, which can be employed as useful priors for designing image denoising methods. Based on the facts that natural images will be sparsely distributed in some transformed domain, wavelet \cite{bayesshrink} and curvelet \cite{curvelet} transforms have been widely adopted for image denoising. The sparse representation based methods \cite{ksvd,lssc,ncsr,bm3d,cbm3d,zhou2012nonparametric} encode image patches over a dictionary by using $\ell_{1}$-norm minimization to enforce the sparsity. The well-known bilateral filters \cite{Tomasi1998} employ the prior information that image pixels exhibit similarity in both spatial domain and intensity domain. Other image priors such as multiscale self-similarity \cite{blsgsm} and nonlocal self-similarity \cite{nlm,nlbayes}, or the combination of multiple image priors \cite{wnnm,pgpd} have also been successfully used in image denoising. For example, by using low-rank minimization to characterize the image nonlocal self-similarity, the WNNM \cite{wnnm} method achieves state-of-the-art performance for AWGN denoising. 

Instead of using predefined image priors, methods have also been proposed to learn priors from natural images for denoising. The generative image prior learning methods usually learn prior models from a set of external clean images and apply the learned prior models to the given noisy image \cite{foe,epll,pgpd,combexin,external,tid}, or learn priors from the given noisy image to perform denoising \cite{ksvd}. Recently, the discriminative image prior learning methods \cite{mlp,xie2012image,dncnn,
chen2015learning,barbu2009training,csf}, which learn denoising models from pairs of clean and noisy images, have been becoming popular. The representative methods include the neural network based methods \cite{mlp,xie2012image,dncnn}, random fields based methods \cite{barbu2009training,csf}, and reaction diffusion based methods \cite{chen2015learning}.

\begin{figure*}
\vspace{-3mm}
\centering
\subfigure{
\begin{minipage}[t]{0.19\textwidth}
\centering
\raisebox{-0.15cm}{\includegraphics[width=1\textwidth]{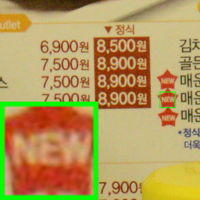}}
{\footnotesize (a) Noisy \cite{crosschannel2016}: 33.30dB }
\end{minipage}
\begin{minipage}[t]{0.19\textwidth}
\centering
\raisebox{-0.15cm}{\includegraphics[width=1\textwidth]{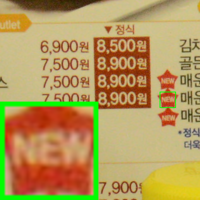}}
{\footnotesize (b) CBM3D \cite{cbm3d}: 34.55dB  }
\end{minipage}
\begin{minipage}[t]{0.19\textwidth}
\centering
\raisebox{-0.15cm}{\includegraphics[width=1\textwidth]{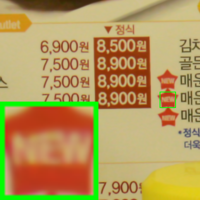}}
{\footnotesize (c) WNNM \cite{wnnm}: 35.85dB  }
\end{minipage}
\begin{minipage}[t]{0.19\textwidth}
\centering
\raisebox{-0.15cm}{\includegraphics[width=1\textwidth]{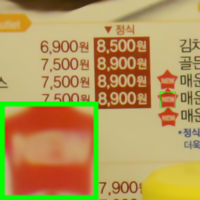}}
{\footnotesize (d) CSF \cite{csf}: 35.39dB }
\end{minipage}
\begin{minipage}[t]{0.19\textwidth}
\centering
\raisebox{-0.15cm}{\includegraphics[width=1\textwidth]{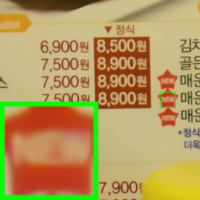}}
{\footnotesize (e) TNRD \cite{chen2015learning}: 35.97dB   }
\end{minipage}
}\vspace{-3mm}
\subfigure{
\begin{minipage}[t]{0.19\textwidth}
\centering
\raisebox{-0.15cm}{\includegraphics[width=1\textwidth]{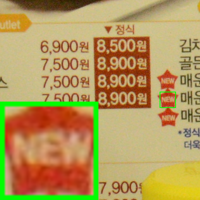}}
{\footnotesize (f) DnCNN \cite{dncnn}: 34.14dB }
\end{minipage}
\begin{minipage}[t]{0.19\textwidth}
\centering
\raisebox{-0.15cm}{\includegraphics[width=1\textwidth]{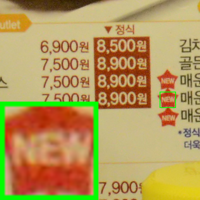}}
{\footnotesize (g) NI \cite{neatimage}: 34.39dB  }
\end{minipage}
\begin{minipage}[t]{0.19\textwidth}
\centering
\raisebox{-0.15cm}{\includegraphics[width=1\textwidth]{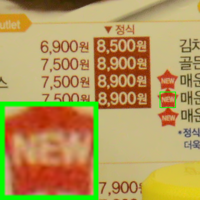}}
{\footnotesize (h) NC \cite{noiseclinic,ncwebsite}: 35.33dB   }
\end{minipage}
\begin{minipage}[t]{0.19\textwidth}
\centering
\raisebox{-0.15cm}{\includegraphics[width=1\textwidth]{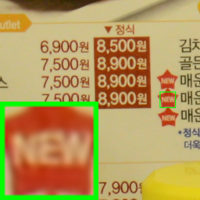}}
{\footnotesize (i) Ours: \textbf{37.49}dB  }
\end{minipage}
\begin{minipage}[t]{0.19\textwidth}
\centering
\raisebox{-0.15cm}{\includegraphics[width=1\textwidth]{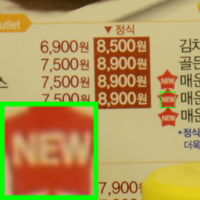}}
{\footnotesize (j) Mean Image \cite{crosschannel2016} }
\end{minipage}
}\vspace{-3mm}
\caption{Denoised images of a region cropped from the real-world noisy image ``Nikon D800 ISO 3200 A3" \cite{crosschannel2016} by different methods.\ The scene was shot 500 times with the same camera and camera setting.\ The mean image of the 500 shots is roughly taken as the ``ground truth", with which the PSNR can be computed. The images are better viewed by zooming in on screen.} 
\label{fig1}
\vspace{-4mm}
\end{figure*}

Most of the above mentioned methods focus on AWGN removal, however, the assumption of AWGN is too ideal to be true for real-world noisy images, where the noise is much more complex and varies with different scenes, cameras and camera settings (ISO, shutter speed, and aperture, etc.) \cite{crosschannel2016,healey1994radiometric}. As a result, many denoising methods in literature, including those learning based methods, become less effective when applied to real-world noisy images. Fig. \ref{fig1} shows an example, where we apply some representative and state-of-the-art denoising methods, including CBM3D \cite{cbm3d}, WNNM \cite{wnnm}, DnCNN \cite{dncnn}, CSF \cite{csf}, and TNRD \cite{chen2015learning} to a real-world noisy image (captured by a Nikon D800 camera with ISO is 3200) provided in \cite{crosschannel2016}. One can see that these methods either remain much the noise or over-smooth the image details.

There have been a few methods \cite{fullyblind,rabie2005robust,Liu2008,almapg,crosschannel2016,Zhu_2016_CVPR,
noiseclinic,ncwebsite,mcwnnm} and software toolboxes \cite{neatimage} developed for real-world noisy image denoising. Almost all of these methods follow a two-stage framework: first estimate the parameters of the noise model (usually assumed to be Gaussian or mixture of Gaussians (MoG)), and then perform denoising with the estimated noise model. However, the noise in real-world noisy images is very complex and is hard to be modeled by explicit distributions such as Gaussian and MoG. According to \cite{healey1994radiometric}, the noise corrupted in the in-camera imaging process \cite{tsin2001statistical,NewInCamera,crosschannel2016,karaimer_brown_ECCV_2016} is signal dependent and comes from five main sources: photon shot, fixed pattern, dark current, readout, and quantization noise. The existing methods \cite{fullyblind,rabie2005robust,Liu2008,almapg,crosschannel2016,Zhu_2016_CVPR,noiseclinic,
ncwebsite,neatimage} mentioned above may not perform well on real-world noisy image denoising tasks. Fig. \ref{fig1} also shows the denoising results of two real-world noisy image denoising methods, Noise Clinic \cite{noiseclinic,ncwebsite} and Neat Image \cite{neatimage}. One can see that these two methods still generate much noise caused artifacts. 

\begin{figure*}
\vspace{-4mm}
\centering
\captionsetup{justification=centering,margin=0.1cm}
\includegraphics[width=0.65\linewidth]{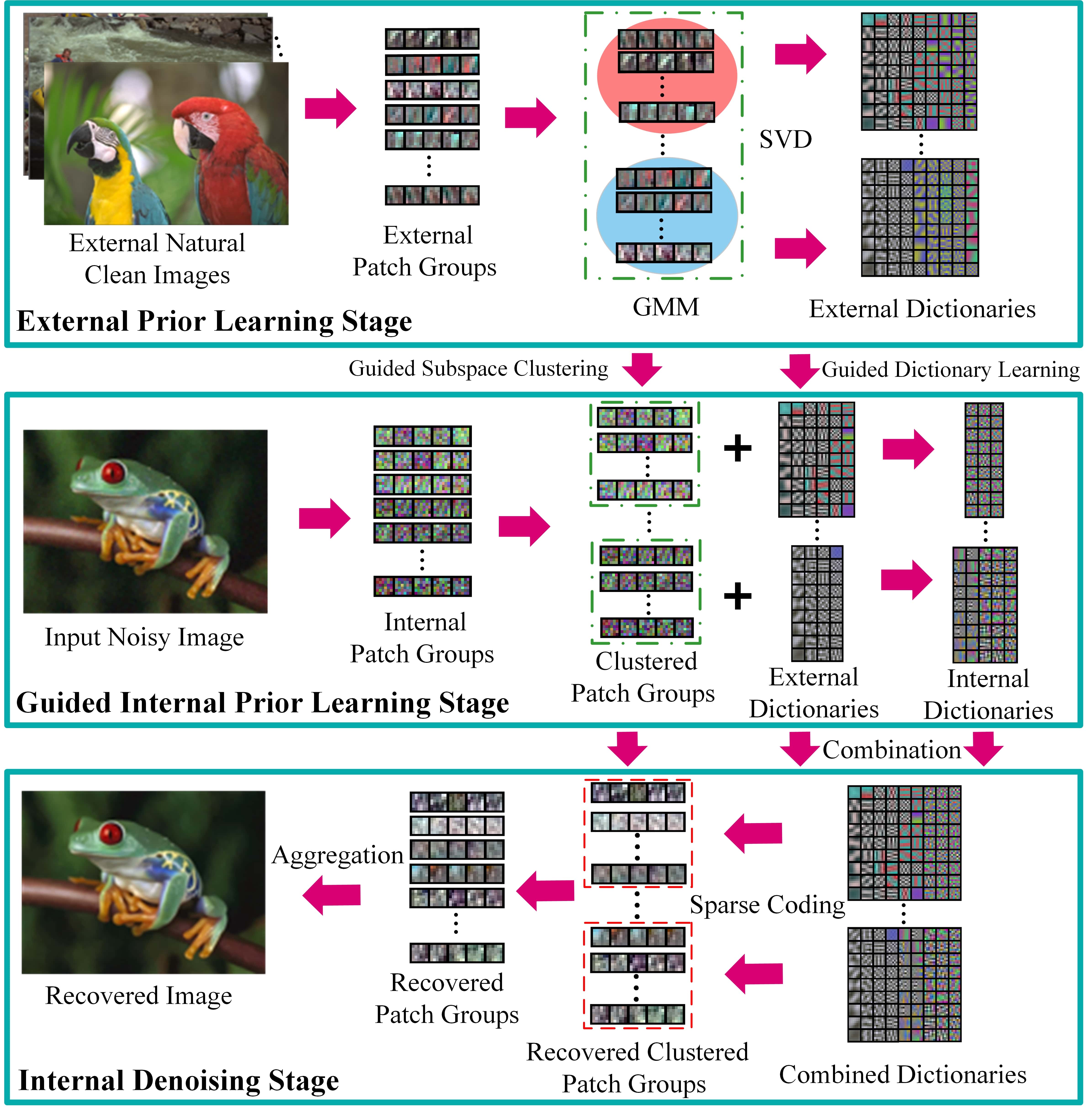}
\vspace{-2mm}
\centering
\caption{Flowchart of the proposed external prior guided internal prior learning and denoising framework.}
\label{fig2}
\vspace{-4mm}
\end{figure*}

This work aims to develop a new paradigm for real-world noisy image denoising. Different from existing real-world noisy image denoising methods \cite{fullyblind,rabie2005robust,Liu2008,almapg,crosschannel2016,Zhu_2016_CVPR,noiseclinic,ncwebsite}  which focus on noise modeling, we focus on image prior learning.\ We argue that with a strong and adaptive prior learning scheme, robust denoising performance on real-world noisy images can still be obtained.\ To achieve this goal, we propose to first learn image priors from external clean images, and then employ the learned external priors to guide the learning of internal priors from the given noisy image.\ The flowchart of the proposed method is illustrated in Fig.\ \ref{fig2}.\ We first extract millions of patch groups (PGs) from a set of high quality natural images, with which a Gaussian Mixture Model (GMM) is learned as the external image prior. The learned GMM prior model is used to assign each PG extracted from the given noisy image into its most suitable cluster via maximum a-posterior, and then an external-internal hybrid orthogonal dictionary is learned as the final prior for each cluster, with which the denoising can be readily performed by weighted sparse coding with closed form solution. The external priors learned from clean images preserve fine-scale image structural information, which is hard to be reproduced from noisy images. Therefore, external dictionary can serve as a good supplement to the internal dictionary. Our proposed denoising method is simple and efficient, yet our extensive experiments on real-world noisy images demonstrate its better denoising performance than the current state-of-the-arts.

\vspace{-2mm}
\section{Related Work}

\vspace{-1mm}
\subsection{Internal and External Prior Learning}
\vspace{-1mm}

Learning natural image priors plays a key role in image denoising
\cite{blsgsm,zhou2012nonparametric,ksvd,lssc,ncsr,foe,epll,pgpd,combexin,external,
tid,mlp,xie2012image,dncnn,barbu2009training,csf,chen2015learning}. There are mainly four categories of prior learning based methods. 1) External prior learning methods \cite{foe,epll,pgpd} learn priors (e.g., dictionaries) from a set of external clean images, and the learned priors are used to recover the latent clean image from the given noisy image. 2) Internal prior learning methods \cite{blsgsm,zhou2012nonparametric,ksvd,lssc,ncsr} directly learn priors from a given noisy image, and image denoising is often done simultaneously with the prior learning process. 3) Discriminative prior learning methods \cite{mlp,xie2012image,dncnn,barbu2009training,
csf,chen2015learning} learn discriminative models or mapping functions from clean and noisy image pairs, and the learned models or mapping functions are applied to a noisy image for denoising. 4) Hybrid methods \cite{combexin,external,tid} combine the external and internal priors to denoise the given input image.

It has been shown \cite{foe,epll,pgpd} that the external priors learned from natural clean images are effective and efficient for universal image denoising problems, whereas they are not adaptive to the given noisy image and some fine-scale image structures may not be well recovered. By contrast, the internal priors learned from the given noisy image are adaptive to image content, but the learned priors can be much affected by noise and the learning processing is usually slow \cite{blsgsm,zhou2012nonparametric,ksvd,lssc,ncsr}. Besides, most of the internal prior learning methods \cite{blsgsm,zhou2012nonparametric,ksvd,lssc,ncsr} assume additive white Gaussian noise (AWGN), making the learned priors less robust for real-world noisy images. In this paper, we use external priors to guide the internal prior learning. Our method is not only much faster than the traditional internal learning methods, but also very robust to denoise real-world noisy images.

In \cite{combexin}, the authors employed external clean patches to denoise noisy patches with high individual Signal-to-Noise-Ratio (PatchSNR), and employed internal noisy patches to denoise noisy patches with low PatchSNR. This is essentially different from our work which employs the external patch group based prior to guide the clustering and dictionary learning of the internal noisy patch groups. In \cite{external}, the external priors are only used to guide the internal patch clustering for image denoising, while in our work, the learned external priors are employed to guide not only the internal clustering, but also the internal dictionary learning. Besides, the method of \cite{external} follows a patch based framework for AWGN removal, while in our work we employ a patch group based framework for real-world noisy image denoising. In addition, some technical details are also different. For example, method in \cite{external} utilizes low-rank minimization for denoising, while we use dictionary learning and sparse coding for denoising. In the Targeted Image Denoising (TID) method \cite{tid}, targeted images are selected from a large dataset for each patch in the input noisy image for denoising, which is computationally expensive.

\vspace{-3mm}
\subsection{Real-World Noisy Image Denoising}
\vspace{-1mm}

Most of the denoising methods in literature \cite{bayesshrink,curvelet,ksvd,lssc,ncsr,bm3d,cbm3d,
zhou2012nonparametric,Tomasi1998,blsgsm,nlm,nlbayes,wnnm,pgpd,foe,epll,
mlp,xie2012image,dncnn,barbu2009training,csf,chen2015learning} assume AWGN noise and use simulated noisy images for algorithm design and evaluation. Recently, several denoising methods have been proposed to remove unknown noise from real-world noisy images \cite{fullyblind,rabie2005robust,Liu2008,almapg,crosschannel2016,Zhu_2016_CVPR,noiseclinic,ncwebsite}. Portilla \cite{fullyblind} employed a correlated Gaussian model to estimate the noise of each wavelet subband. Rabie \cite{rabie2005robust} modeled the noisy pixels as outliers and performed denoising via Lorentzian robust estimator. Liu et al. \cite{Liu2008} proposed the ``noise level function'' to estimate the noise and performed denoising by learning a Gaussian conditional random field. Gong et al. \cite{almapg} proposed to model the data fitting term via weighted sum of $\ell_{1}$ and $\ell_{2}$ norms and performed denoising by a simple sparsity regularization term in the wavelet transform domain. The ``Noise Clinic'' \cite{noiseclinic,ncwebsite} estimates the noise distribution by using a multivariate Gaussian model and removes the noise by using a generalized version of nonlocal Bayesian model \cite{nlbayes}. Zhu et al. \cite{Zhu_2016_CVPR} proposed a Bayesian method to approximate and remove the noise via a low-rank mixture of Gaussians (MoG) model. The method in \cite{crosschannel2016} models the cross-channel noise in real-world noisy image as a multivariate Gaussian and the noise is removed by the Bayesian nonlocal means filter \cite{kervrann2007bayesian}. The commercial software Neat Image \cite{neatimage} estimates the noise parameters from a flat region of the given noisy image and filters the noise correspondingly.

The methods \cite{fullyblind,rabie2005robust,Liu2008,almapg,crosschannel2016,Zhu_2016_CVPR,noiseclinic,ncwebsite}  emphasize much on the noise modeling, and they use Gaussian or MoG to model the noise in real-world noisy images. Nonetheless, the noise in real-world noisy images is very complex and hard to be modeled by explicit distributions \cite{healey1994radiometric}. These works ignore the importance of learning image priors, which actually can be easier to model compared with modeling the complex realistic noise. In this paper, we propose a simple yet effective image prior learning method for real-world noisy image denoising. Due to its strong prior modeling ability, the proposed method simply models the noise as locally Gaussian, and it achieves highly competitive performance on real-world noisy image denoising.

\vspace{-1mm}
\section{External Prior Guided Internal Prior Learning for Image Denoising}
\vspace{-1mm}

In this section, we first describe the learning of external prior, and then describe in detail the guided internal prior learning method, followed by the denoising algorithm.

\vspace{-1mm}
\subsection{Learn External Patch Group Priors}
\vspace{-1mm}

The nonlocal self-similarity based patch group (PG) prior learning \cite{pgpd} has proved to be very effective for image denosing. In this work, we extract PGs from natural clean images to learn external priors. A PG is a group of similar patches to a local patch. In our method, each local patch is extracted from a RGB image with patch size $p\times p \times 3$.\ We search the $M$ most similar (i.e., smallest Euclidean distance) patches to this local patch (including the local patch itself) in a $W\times W$ region around it. Each patch is stretched to a patch vector $\mathbf{x}_{m}\in \mathbb{R}^{3p^{2}\times1}$ to form the PG, denoted by $\{\mathbf{x}_{m}\}_{m=1}^{M}$.\ The mean vector of this PG is $\bm{\mu}=\frac{1}{M}\sum_{m=1}^{M}\mathbf{x}_{m}$, and the group mean subtracted PG is defined as $\mathbf{\overline{X}}\triangleq \{\mathbf{\overline{x}}_{m}=\mathbf{x}_{m}-\bm{\mu}\}_{m=1}^{M}$.

Assume that a number of $L$ PGs are extracted from a set of external natural images, and the $l$-th PG is $\mathbf{\overline{X}}_{l}\triangleq \{\mathbf{\overline{x}}_{l,m}\}_{m=1}^{M}, l=1,...,L$.\ A Gaussian Mixture Model (GMM) is learned to model the PG prior.\ The overall log-likelihood function is
\vspace{-2mm}
\begin{equation}\label{equ1}
\vspace{-2mm}
\begin{split}
\ln\mathcal{L}=\sum_{l=1}^{L} \ln(\sum_{k=1}^{K}\pi_{k}\prod_{m=1}^{M}\mathcal{N}(\mathbf{\overline{x}}_{l,m}|\bm{\mu}_{k},\bm{\Sigma}_{k})).
\end{split}
\end{equation}
The learning process is similar to the GMM learning in \cite{pgpd,epll,partial}.\ Finally, a GMM model with $K$ Gaussian components is learned, and the learned parameters include mixture weights $\{\pi_{k}\}_{k=1}^{K}$, mean vectors $\{\bm{\mu}_{k}\}_{k=1}^{K}$, and covariance matrices $\{\bm{\Sigma}_{k}\}_{k=1}^{K}$.\ Note that the mean vector of each cluster is naturally zero, i.e., $\bm{\mu}_{k}=\bm{0}$.  

To better describe the subspace of each Gaussian component, we perform singular value decomposition (SVD) \cite{eckart1936approximation} on the covariance matrix:
\vspace{-2mm}
\begin{equation}\label{equ2}
\vspace{-2mm}
\bm{\Sigma}_{k}=\bm{U}_{k}\bm{S}_{k}\bm{U}_{k}^{\top}.
\end{equation}
The eigenvector matrices $\{\bm{U}_{k}\}_{k=1}^{K}$ will be employed as the external orthogonal dictionary to guide the internal sub-dictionary learning in next sub-section.\ The singular values in $\bm{S}_{k}$ reflect the significance of the singular vectors in $\bm{U}_{k}$.\ They  will also be utilized as prior weights for weighted sparse coding in our denoising algorithm.

\subsection{Guided Internal Prior Learning}

After the external PG prior model is learned from external natural clean images, we employ it to guide the internal PG prior learning for a given real-world noisy image.\ The guidance lies in two aspects. First, the external prior will guide the subspace clustering \cite{vidalsc,rssc} of internal noisy PGs. Second, the external prior will guide the orthogonal dictionary learning of internal noisy PGs.

\subsubsection{Internal Subspace Clustering}

Given a real-world noisy image $\mathbf{y}$, we extract $N$ (overlapped) local patches from it.\ Similar to the external prior learning stage, for the $n$-th ($n=1,...,N$) local patch we search its $M$ most similar (by Euclidean distance) patches around it to form a noisy PG, denoted by $\bm{Y}_{n} = \{\mathbf{y}_{n,1},...,\mathbf{y}_{n,M}\}$.\ Then the group mean of $\bm{Y}_{n}$, denoted by $\bm{\mu}_{n}$, is subtracted from each patch by $\bm{\overline{y}}_{n,m}\triangleq\mathbf{y}_{n,m}-\bm{\mu}_{n}$, leading to the mean subtracted noisy PG $\bm{\overline{Y}}_{n}\triangleq \{\bm{\overline{y}}_{n,m}\}_{m=1}^{M}$.

The external GMM prior models $\{\mathcal{N}(\bm{0},\bm{\Sigma}_{k})\}_{k=1}^{K}$ basically characterize the subspaces of natural high quality PGs.\ Therefore, we project each noisy PG $\bm{\overline{Y}}_{n}$ into the subspaces of $\{\mathcal{N}(\bm{0},\bm{\Sigma}_{k})\}_{k=1}^{K}$ and assign it to the most suitable subspace based on the posterior probability:
\vspace{-2mm}
\begin{equation}\label{equ3}
\vspace{-2mm}
P(k|\bm{\overline{Y}}_{n})=\frac{\prod_{m=1}^{M}\mathcal{N}(\bm{\overline{y}}_{n,m}|\bm{0},\bm{\Sigma}_{k})}{\sum_{l=1}^{K}\prod_{m=1}^{M}\mathcal{N}(\bm{\overline{y}}_{n,m}|\bm{0},\bm{\Sigma}_{l})}
\end{equation}
for $k=1,...,K$.\ Then $\bm{\overline{Y}}_{n}$ is assigned to the subspace with the maximum \emph{a}-posteriori (MAP) probability $\max_{k}P(k|\bm{\overline{Y}}_{n})$.

\subsubsection{Guided Orthogonal Dictionary Learning}

Assume that we have assigned all the internal noisy PGs $\{\bm{\overline{Y}}_{n}\}_{n=1}^{N}$ to their corresponding most suitable subspaces in $\{\mathcal{N}(\bm{0},\bm{\Sigma}_{k})\}_{k=1}^{K}$. For the $k$-th subspace, the noisy PGs assigned to it are $\{\bm{\overline{Y}}_{k_{n}}\}_{n=1}^{N_{k}}$, where $\bm{\overline{Y}}_{k_{n}}=[\bm{\overline{y}}_{k_{n},1},...,\bm{\overline{y}}_{k_{n},M}]$ and $\sum_{k=1}^{K}N_{k}=N$.\ We propose to learn an orthogonal dictionary $\bm{D}_{k}$ from each set of PGs $\bm{\overline{Y}}_{k_{n}}$ to characterize the internal PG prior with the guidance of the corresponding external orthogonal dictionary $\bm{U}_{k}$ (Eq.\ (\ref{equ2})). The reasons that we learn orthogonal dictionaries are two-fold.\ Firstly, the PGs $\{\bm{\overline{Y}}_{k_{n}}\}_{n=1}^{N_{k}}$ are in a subspace of the whole space of all PGs; therefore, there is no necessary to learn a redundant over-complete dictionary to characterize it, while an orthonormal dictionary has naturally zero \emph{mutual incoherence} \cite{donoho2001uncertainty}. Secondly, the orthogonality of dictionary can make the patch encoding in the testing stage very efficient, leading to an efficient denoising algorithm (please refer to sub-section III-C for more details).

We let the orthogonal dictionary $\bm{D}_{k}$ be 
\vspace{-2mm}
\begin{equation}
\vspace{-2mm}
\label{equ4}
\bm{D}_{k}\triangleq[\bm{D}_{k,\text{E}}\ \bm{D}_{k,\text{I}}]\in \mathbb{R}^{3p^2\times 3p^2},
\end{equation}
where $\bm{D}_{k,\text{E}}=\bm{U}_{k}(:,1:r)\in\mathbb{R}^{3p^2\times r}$ is the external sub-dictionary and it includes the first $r$ most important eigenvectors of $\bm{U}_{k}$, and the internal sub-dictionary $\bm{D}_{k,\text{I}}\in\mathbb{R}^{3p^2\times (3p^2-r)}$ is to be adaptively learned from the noisy PGs $\{\bm{\overline{Y}}_{k_{n}}\}_{n=1}^{N_{k}}$.\ The rationale to design $\bm{D}_{k}$ as a hybrid dictionary is as follows.\ The external sub-dictionary $\bm{D}_{k,\text{E}}$ is pre-trained from external clean data, and it represents the $k$-th latent subspace of natural images, which is helpful to reconstruct the common latent structures of images. However, $\bm{D}_{k,\text{E}}$ is general to all images but not adaptive to the given noisy image.\ Some fine-scale details specific to the given image may not be well characterized by $\bm{D}_{k,\text{E}}$. Therefore, we learn an internal sub-dictionary $\bm{D}_{k,\text{I}}$ to supplement $\bm{D}_{k,\text{E}}$.\ In other words, $\bm{D}_{k,\text{I}}$ is to reveal the latent subspace adaptive to the input noisy image, which cannot be effectively represented by $\bm{D}_{k,\text{E}}$. 

For notation simplicity, in the following development we ignore the subspace index $k$ for $\bm{\overline{Y}}_{k_{n}}$ and $\bm{D}_{k}$, etc.\ The learning of hybrid orthogonal dictionary $\bm{D}$ is performed under the following weighted sparse coding
framework:
\vspace{-2mm}
\begin{equation}\label{equ5}
\vspace{-2mm}
\begin{split}
\min_{\bm{D}_{\text{I}},\{\bm{\alpha}_{n,m}\}}
&\sum_{n=1}^{N}\sum_{m=1}^{M}(\|\bm{\overline{y}}_{n,m}-\bm{D}\bm{\alpha}_{n,m}\|_{2}^{2}+\sum_{j=1}^{3p^{2}}\lambda_{j}|\bm{\alpha}_{n,m,j}|)
\\
&
\text{s.t.}
\quad
\bm{D}=[\bm{D}_{\text{E}}\ \bm{D}_{\text{I}}],\ \bm{D}^{\top}\bm{D} = \bm{I},
\end{split}
\end{equation}
where $\bm{I}$ is the $3p^{2}$ dimensional identity matrix, $\bm{\alpha}_{n,m}$ is the sparse coding vector of the $m$-th patch $\bm{\overline{y}}_{n,m}$ in the $n$-th PG $\bm{\overline{Y}}_{n}$ and $\bm{\alpha}_{n,m,j}$ is the $j$-th element of $\bm{\alpha}_{n,m}$. $\lambda_{j}$ is the $j$-th regularization parameter defined as
\vspace{-2mm}
\begin{equation}\label{equ6}
\vspace{-2mm}
\lambda_{j} = \lambda/(\sqrt{\bm{S}_{k}(j)}+\varepsilon),
\end{equation}
where $\bm{S}_{k}(j)$ is the $j$-th singular value of diagonal singular value matrix $\bm{S}_{k}$ (please refer to Eq.\ (\ref{equ2})) and $\varepsilon$ is a small positive number to avoid zero denominator.\ Note that $\bm{D}_{\text{E}}=\bm{U}_{k}$ if $r=3p^{2}$ and $\bm{D}_{\text{E}}=\emptyset$ if $r=0$.

In the dictionary learning model (\ref{equ5}), we use the $\ell_{2}$ norm to model the representation residual of PGs. This is because the patches in those PGs have similar content, and we assume that the noise therein will have similar statistics, which can be roughly modeled as locally Gaussian. On the other hand, this will make the dictionary learning much easier to solve. We employ an alternating iterative approach to solve the optimization problem (\ref{equ5}). Specifically, we initialize the orthogonal dictionary as $\bm{D}^{(0)}=\bm{U}_{k}$ and for $t=0,1, ...,T-1$, and alternatively update $\bm{\alpha}_{n,m}$ and $\bm{D}_{\text{I}}$ as follows.

\vspace{2mm}
\textbf{Updating Sparse Coding Coefficients}: Given the orthogonal dictionary $\bm{D}^{(t)}$, we update each sparse coding vector $\bm{\alpha}_{n,m}$ by solving
\vspace{-2mm}
\begin{equation}\label{equ7}
\vspace{-2mm}
\begin{split}
\bm{\alpha}_{n,m}^{(t+1)}:=\argmin_{\bm{\alpha}_{n,m}}
\|\bm{\overline{y}}_{n,m}-\bm{D}^{(t)}\bm{\alpha}_{n,m}\|_{2}^{2}+\sum_{j=1}^{3p^{2}}\lambda_{j}|\bm{\alpha}_{n,m,j}|.
\end{split}
\end{equation}
Since dictionary $\bm{D}^{(t)}$ is orthogonal, the problems (\ref{equ7}) has a closed-form solution
\vspace{-2mm}
\begin{equation}\label{equ8}
\bm{\alpha}_{n,m}^{(t+1)}= \text{sgn}((\bm{D}^{(t)})^{\top}\bm{\overline{y}}_{n,m})\odot \text{max}(|(\bm{D}^{(t)})^{\top}\bm{\overline{y}}_{n,m}|-\bm{\lambda},\bm{0}),
\end{equation}
where $\bm{\lambda} = \frac{1}{2}[\lambda_{1},\lambda_{2},...,\lambda_{3p^2}]^{\top}$ is the vector of regularization parameter, $\text{sgn}(\bullet)$ is the sign function and $\odot$ means element-wise multiplication.\ The detailed derivation of Eq. (\ref{equ8}) can be found in Appendix A.
 
\vspace{2mm}
\textbf{Updating Internal Sub-dictionary}: Given the sparse coding vectors $\{\bm{\alpha}_{n,m}^{(t+1)}\}$, we update the internal sub-dictionary by solving
\vspace{-2mm}
\begin{equation}\label{equ9}
\begin{split}
\bm{D}_{\text{I}}^{(t+1)}
:
&
=
\argmin_{\textbf{D}_{\text{I}}}
\sum_{n=1}^{N}\sum_{m=1}^{M}\|\bm{\overline{y}}_{n,m}-\bm{D}\bm{\alpha}_{n,m}^{(t+1)}\|_{2}^{2}
\\
&
=
\argmin_{\textbf{D}_{\text{I}}}
\|\bm{\overline{Y}}_{n}-\bm{D}\bm{A}^{(t+1)}\|_{F}^{2}
\\
\text{s.t.}
\quad
\bm{D}
&
=
[\bm{D}_{\text{E}}\ \bm{D}_{\text{I}}],\ \bm{D}_{\text{I}}^{\top}\bm{D}_{\text{I}} = \bm{I}_{(3p^2-r)},\ \bm{D}_{\text{E}}^{\top}\bm{D}_{\text{I}} = \bm{0},
\end{split}
\end{equation}
where $\textbf{A}^{(t+1)}=[\bm{\alpha}_{1,1}^{(t+1)},...,\bm{\alpha}_{1,M}^{(t)},...,\bm{\alpha}_{N,1}^{(t+1)},...,\bm{\alpha}_{N,M}^{(t+1)}]$ and $\bm{I}_{(3p^2-r)}$ is the $(3p^2-r)$ dimensional identity matrix.\ The sparse coefficients matrix can be written as $\bm{A}^{(t+1)}=[(\bm{A}_{\text{E}}^{(t+1)})^{\top}\ (\bm{A}_{\text{I}}^{(t+1)})^{\top}]^{\top}$ where the external part $\bm{A}_{\text{E}}^{(t+1)}\in\mathbb{R}^{r\times NM}$ and the internal part $\bm{A}_{\text{I}}^{(t+1)}\in\mathbb{R}^{(3p^2-r)\times NM}$ represent the coding coefficients of $\bm{Y}$ over external sub-dictionary $\bm{D}_{\text{E}}$ and internal sub-dictionary $\bm{D}_{\text{I}}^{(t)}$, respectively.\ According to the following Theorem \ref{th1}, by setting $\mathcal{Y}=\bm{\overline{Y}}_{n}-\bm{D}_{\text{E}}\bm{A}_{\text{E}}^{(t+1)},\mathcal{E}=\bm{D}_{\text{E}},\mathcal{D}=\bm{D}_{\text{I}},\mathcal{A}=\bm{A}_{\text{I}}$, the problem (\ref{equ9}) has a closed-form solution $\bm{D}_{\text{I}}^{(t+1)}=\bm{U}_{\text{I}}\bm{V}_{\text{I}}^{\top}$, where $\bm{U}_{\text{I}}\in\mathbb{R}^{3p^2\times (3p^2-r)}$ and $\bm{V}_{\text{I}}\in\mathbb{R}^{(3p^2-r)\times (3p^2-r)}$ are the orthogonal matrices obtained by the following SVD \cite{eckart1936approximation}
\vspace{-1mm}
\begin{equation}\label{equ10}
\vspace{-1mm}
(\bm{I}-\bm{D}_{\text{E}}\bm{D}_{\text{E}}^{\top})\bm{Y}(\bm{A}_{\text{I}}^{(t+1)})^{\top}
=
\bm{U}_{\text{I}}\bm{S}_{\text{I}}\bm{V}_{\text{I}}^{\top}.
\end{equation}
The orthogonality of internal sub-dictionary $\bm{D}_{\text{I}}^{(t+1)}$ can be checked by 
$(\bm{D}_{\text{I}}^{(t+1)})^{\top}(\bm{D}_{\text{I}}^{(t+1)})=\bm{V}_{\text{I}}\bm{U}_{\text{I}}^{\top}\bm{U}_{\text{I}}\bm{V}_{\text{I}}^{\top}=\bm{I}_{(3p^2-r)}$.\ In fact, the Theorem \ref{th1} provides a sufficient and necessary condition to guarantee the existence of the closed-form solution for the internal sub-dictionary of the problem (\ref{equ9}).

\begin{theorem}
\label{th1}
Let $\mathcal{A}\in \mathbb{R}^{(3p^2-r)\times M}$, $\mathcal{Y}\in \mathbb{R}^{3p^2\times M}$ be two given data matrices. $\mathcal{E}\in\mathbb{R}^{3p^2\times r}$ is a given matrix satisfying $\mathcal{E}^{\top}\mathcal{E}=\bm{I}_{r\times r}$, then $\hat{\mathcal{D}} = \mathcal{U}\mathcal{V}^{\top}$ is the necessary condition of
\begin{equation}\label{equ11}
\begin{split}
\hat{\mathcal{D}}
=
&
\arg\min_{\mathcal{D}}\|\mathcal{Y}-\mathcal{D}\mathcal{A}\|_{F}^{2}
\quad
\\
&
\text{s.t.}
\quad
\mathcal{D}^{\top}\mathcal{D} = \bm{I}_{(3p^2-r)\times (3p^2-r)}, \mathcal{E}^{\top}\mathcal{D} = \bm{0}_{r\times (3p^2-r)}
,
\end{split}
\end{equation}
where $\mathcal{U}\in \mathbb{R}^{3p^2\times (3p^2-r)}$ and $\mathcal{V}\in \mathbb{R}^{(3p^2-r)\times (3p^2-r)}$ are the orthogonal matrices obtained by performing economy (a.k.a. reduced) SVD  \cite{eckart1936approximation}:
\begin{equation}\label{equ12}
(\bm{I}_{3p^2\times 3p^2}-\mathcal{E}\mathcal{E}^{\top})\mathcal{Y}\mathcal{A}^{\top} = \mathcal{U}\Sigma\mathcal{V}^{\top}
\end{equation}
Besides, if $\text{rank}(\Sigma)=3p^2-r$, $\hat{\mathcal{D}} = \mathcal{U}\mathcal{V}^{\top}$ is also the sufficient condition of problem (\ref{equ11}). 
\end{theorem}

The proof of the Theorem \ref{th1} can be found in Appendix B. Though the problem (\ref{equ9}) has a closed-form solution by SVD \cite{eckart1936approximation}, the uniqueness of solution cannot be guaranteed since the matrices $(\bm{I}_{3p^2\times 3p^2}-\mathcal{E}\mathcal{E}^{\top})\mathcal{Y}\mathcal{A}^{\top}$ as well as $\mathcal{U}$ and $\mathcal{V}$ may be reduced to matrices of lower rank. Hence, we also analyze the uniqueness of the solution $\hat{\mathcal{D}}$ by the following Theorem \ref{th2}, whose proof can be found in Appendix C.

\begin{theorem}
\label{th2}
(a) If $(\bm{I}_{3p^2\times 3p^2}-\mathcal{E}\mathcal{E}^{\top})\mathcal{Y}\mathcal{A}^{\top}\in\mathbb{R}^{3p^2\times (3p^2-r)}$ is nonsingular, i.e., $\text{rank}(\Sigma)=3p^2-r$, then the solution of $\hat{\mathcal{D}}=\mathcal{U}\mathcal{V}^{\top}$ is unique; (b) If $(\bm{I}_{3p^2\times 3p^2}-\mathcal{E}\mathcal{E}^{\top})\mathcal{Y}\mathcal{A}^{\top}$ is singular, i.e., $0\le\text{rank}(\bm{\Sigma})< 3p^2-r$, then the number of possible solutions of $\hat{\mathcal{D}}$ is $2^{3p^2-r-\text{rank}(\bm{\Sigma})}$ for fixed $\mathcal{U}$ and $\mathcal{V}$.
\end{theorem}

The above alternative updating steps are repeated until the number of iterations exceeds a preset threshold. In each step, the energy value of the objective function (\ref{equ5}) is decreased and we empirically found that the proposed model usually converges in 10 iterations. We summarize the procedures in Algorithm 1.

\begin{table}\label{alg1}
\vspace{-2mm}
\begin{tabular}{l}
\hline
\textbf{Algorithm 1}: External Prior Guided Internal Prior Learning
\\
\hline
\textbf{Input:} Matrices $\bm{\overline{Y}}_{n}$, external sub-dictionary $\bm{D}_{\text{E}}$, parameter vector $\bm{\lambda}$
\\
\textbf{Initialization:} initialize $\bm{D}^{(0)}=\bm{U}_{k}$ by Eq. (\ref{equ2});
\\
\textbf{for} $t=0,1, ...,T-1$ \textbf{do}
\\
1. Update $\bm{\alpha}_{n,m}^{(t+1)}$ by Eq.\ (\ref{equ7});
\\
2. Update $\bm{D}_{\text{I}}^{(t+1)}$ by Eq.\ (\ref{equ9});
\\
\textbf{end for}
\\
\textbf{Output:} Internal orthogonal dictionary $\bm{D}_{\text{I}}^{(T)}$ and sparse codes $\textbf{A}^{(T)}$.
\\
\hline
\end{tabular}
\vspace{-3mm}
\end{table}

\subsection{The Denoising Algorithm}

The denoising of the given noisy image $\mathbf{y}$ can be simultaneously done with the guided internal sub-dictionary learning process. Once we obtain the solutions of sparse coding vectors $\{\hat{\bm{\alpha}}_{n,m}^{(T)}\}$ in Eq.\ (\ref{equ8}) and the orthogonal dictionary $\bm{D}^{(T)} = [\bm{D}_{\text{E}}\ \bm{D}_{\text{I}}^{(T)}]$ in Eq.\ (\ref{equ9}), the latent clean patch $\hat{\mathbf{y}}_{n,m}$ of the $m$-th noisy patch in PG $\bm{Y}_{n}$ is reconstructed as
\vspace{-2mm}
\begin{equation}\label{equ13}
\vspace{-2mm}
\hat{\mathbf{y}}_{n,m}=\bm{D}^{(T)}\hat{\bm{\alpha}}_{n,m}^{(T)}+\bm{\mu}_{n},
\end{equation}
where $\bm{\mu}_{n}$ is the group mean of $\bm{Y}_{n}$. The latent clean image is then reconstructed by aggregating all the reconstructed patches in all PGs.\ We perform the above denoising procedures for several iterations for better denoising outputs.\ The proposed denoising algorithm is summarized in Algorithm 2.

\begin{table}[t!]
\vspace{-2mm}
\label{alg2}
\begin{tabular}{l}
\hline
\textbf{Algorithm 2}: External Prior Guided Internal Prior Learning for 
\\
\quad \quad \quad \quad \quad \quad Real-World Noisy Image Denoising
\\
\hline
\textbf{Input:} Noisy image $\mathbf{y}$, external PG prior GMM model
\\
\textbf{Initialization:} $\hat{\mathbf{x}}^{(0)}=\mathbf{y}$;
\\
\textbf{for} $Ite = 1:IteNum$ \textbf{do}
\\
1. Extracting internal PGs $\{\bm{Y}_{n}\}_{n=1}^{N}$ from $\hat{\mathbf{x}}^{(Ite-1)}$;
\\
\textbf{Guided Internal Subspace Clustering:}
\\
\quad\textbf{for} each PG $\bm{Y}_{n}$ \textbf{do}
\\
2.\quad Calculate group mean $\bm{\mu}_{n}$ and form mean subtracted PG $\bm{\overline{Y}}_{n}$;
\\
3.\quad Subspace clustering via Eq. (\ref{equ3});
\\
\quad\textbf{end for}
\\
\textbf{Guided Internal Orthogonal Dictionary Learning:}
\\
\quad\textbf{for} the PGs in each subspace \textbf{do}
\\
4.\quad External PG prior guided internal orthogonal dictionary learning by
\\
\quad \ \ \ solving Eq. (\ref{equ5});
\\
5.\quad Recover each patch in all PGs via Eq. (\ref{equ13});
\\
\quad\textbf{end for}
\\
6. Aggregate the recovered PGs of all subspaces to form the recovered 
\\
\quad image $\hat{\mathbf{x}}^{(Ite)}$;
\\
\textbf{end for}
\\
\textbf{Output:} The denoised image $\hat{\mathbf{x}}$.
\\
\hline
\end{tabular}
\vspace{-3mm}
\end{table}

\section{Experiments}

\subsection{Implementation Details}

The noise in real-world images is very complex due to the many factors such as sensors, lighting conditions and camera settings. It is difficult to evaluate one algorithm by tuning its parameters for all these different settings. In this work, we fix the parameters of our algorithm and apply it to all the testing datasets, though they were captured by different types of sensors and under different camera settings. The parameters of our method include the patch size $p$, the number of similar patches $M$ in a patch group (PG), the window size $W$ for PG searching, the number of Gaussian components $K$ in GMM, the number of atoms $r$ in the external sub-dictionaries, the sparse regularization parameter $\lambda$, the iteration numbers $T$ for solving problem (\ref{equ5}) and $IteNum$ for Alg. 2.

The performance of our proposed method varies little when we set patch size between $p=6$ and $p=9$, and we fix the patch size as $p=6$ to save computational cost. The search window is fixed to $W = 31$ to balance computational cost and denoising accuracy of the proposed method. The number of patches in a patch group is set as $M = 10$, while using more patches will not bring clear benefits. We learn the external GMM prior with 3.6 million PGs extracted from the Kodak PhotoCD Dataset (\url{http://r0k.us/graphics/kodak/}), which includes 24 high quality color images. The number of Gaussians in GMM is set as $K = 32$, while using more Gaussians can only bring slightly better performance but cost more computational resources. The number of atoms in the external sub-dictionaries affects little the performance when it is set between $r = 27$ and $r = 81$, and we set it as $r = 54$ to make the external and internal sub-dictionaries have the same number of atoms. We set the number of iterations as $T = 2$ for solving the problem (\ref{equ5}), while the number of iterations for Alg. 2 is set as $IteNum = 4$. 

\begin{figure}[t!]
\vspace{-4mm}
\centering
\captionsetup{justification=centering,margin=0.1cm}
\includegraphics[width=1\linewidth]{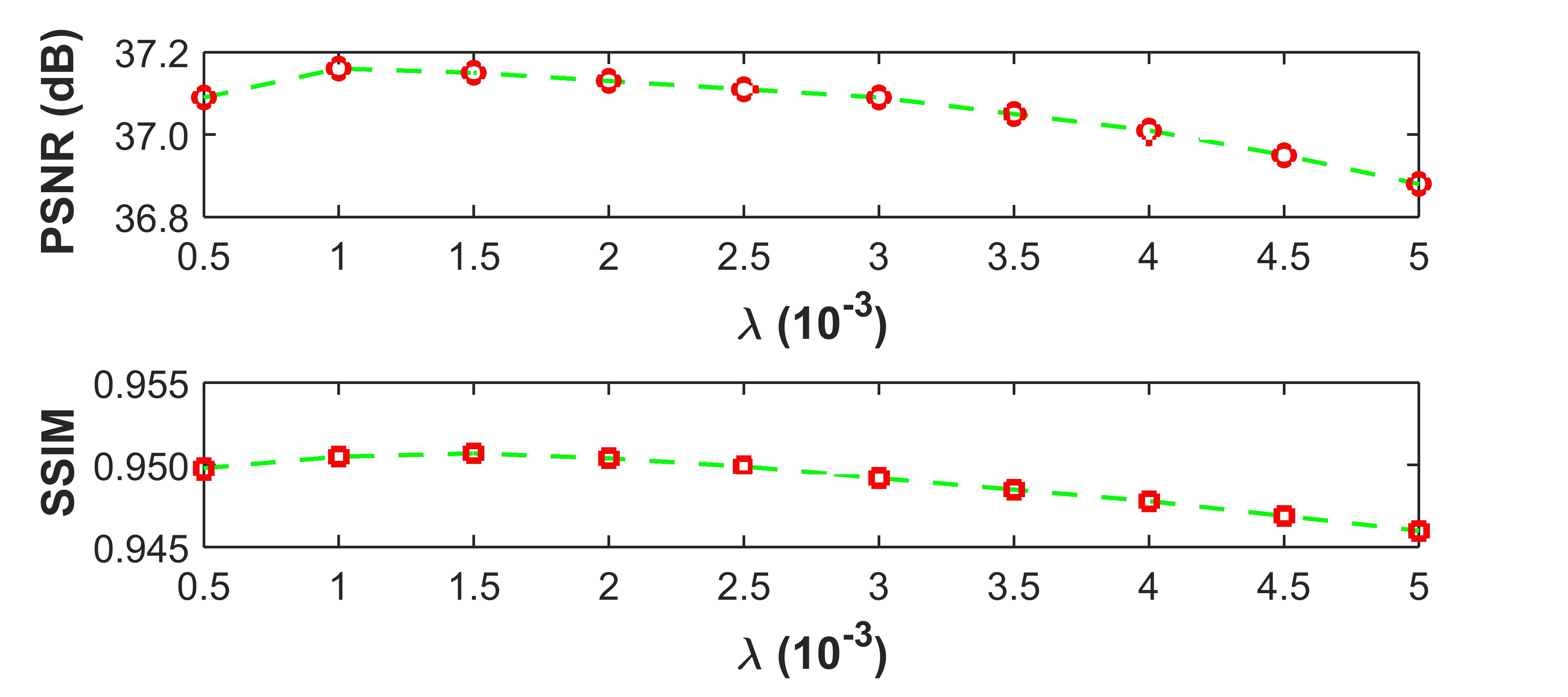}
\vspace{-7mm}
\caption{The influence of parameter $\lambda$ on the average PSNR (dB)/SSIM results of the proposed method on dataset \cite{crosschannel2016}.}
\vspace{-6mm}
\label{fig3}
\end{figure}

One key parameter of our model is the regularization parameter $\lambda$. Fig.\ \ref{fig3} plots the curves of PSNR/SSIM results w.r.t $\lambda$ on the 15 cropped image in dataset \cite{crosschannel2016}. One can see that our proposed method achieves good PSNR/SSIM performance within a certain range of $\lambda$. Similar observations can be made on other datasets. We fix $\lambda=0.001$ in the paper, and it works well across the three datasets used in our experiments. 

All the parameters of our method are fixed in all experiments, which are run under the Matlab2014b environment on a machine with Intel(R) Core(TM) i7-5930K CPU of 3.5GHz and 32GB RAM. We will release the code with the publication of this work.

\subsection{The Testing Datasets}

We evaluate the proposed method on three real-world noisy image datasets, where the images were captured under indoor or outdoor lighting conditions by different types of cameras and camera settings. 

\textbf{Dataset 1.} The first dataset is provided in \cite{crosschannel2016}, which includes noisy images of 11 static scenes.\ The noisy images were collected under controlled indoor environment.\ Each scene was shot 500 times under the same camera and camera setting.\ The mean image of the 500 shots is roughly taken as the ``ground truth", with which the PSNR and SSIM \cite{ssim}can be computed. 

\begin{figure}[t!]
\vspace{-3mm}
\centering
\subfigure{
\begin{minipage}{0.089\textwidth}
\includegraphics[width=1\textwidth]{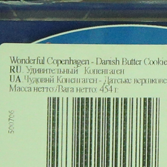}
\end{minipage}
\begin{minipage}{0.089\textwidth}
\includegraphics[width=1\textwidth]{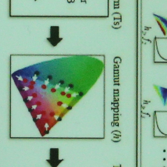}
\end{minipage}
\begin{minipage}{0.089\textwidth}
\includegraphics[width=1\textwidth]{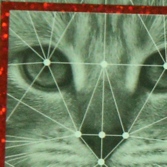}
\end{minipage}
\begin{minipage}{0.089\textwidth}
\includegraphics[width=1\textwidth]{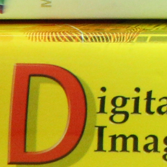}
\end{minipage}
\begin{minipage}{0.089\textwidth}
\includegraphics[width=1\textwidth]{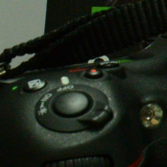}
\end{minipage}
}\vspace{-3mm}
\subfigure{
\begin{minipage}{0.089\textwidth}
\includegraphics[width=1\textwidth]{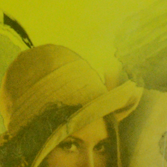}
\end{minipage}
\begin{minipage}{0.089\textwidth}
\includegraphics[width=1\textwidth]{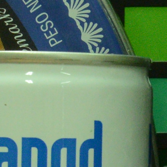}
\end{minipage}
\begin{minipage}{0.089\textwidth}
\includegraphics[width=1\textwidth]{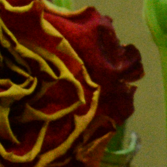}
\end{minipage}
\begin{minipage}{0.089\textwidth}
\includegraphics[width=1\textwidth]{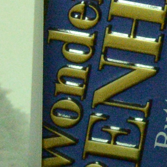}
\end{minipage}
\begin{minipage}{0.089\textwidth}
\includegraphics[width=1\textwidth]{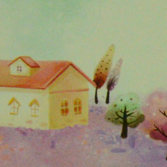}
\end{minipage}
}\vspace{-3mm}
\caption{Some sample images from the \textbf{Dataset 1} \cite{crosschannel2016}.}\vspace{-3mm}
\label{fig4}
\vspace{-3mm}
\end{figure}

Since the image size is very large (about $7000\times5000$) and the 11 scenes share repetitive contents, the authors of \cite{crosschannel2016} cropped 15 smaller images (of size $512\times512$) to perform experiments.\ In order to evaluate the proposed methods more comprehensively, we cropped 60 images of size $500\times500$ from the dataset for experiments. Some samples are shown in Fig.\ \ref{fig4}.\ Note that our cropped 60 images and the 15 cropped images by the authors of \cite{crosschannel2016} are from different shots.

\begin{figure}[ht!]
\vspace{-2mm}
\centering
\subfigure{
\begin{minipage}{0.089\textwidth}
\includegraphics[width=1\textwidth]{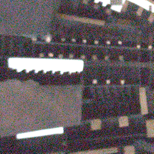}
\end{minipage}
\begin{minipage}{0.089\textwidth}
\includegraphics[width=1\textwidth]{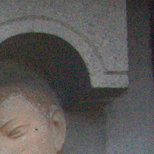}
\end{minipage}
\begin{minipage}{0.089\textwidth}
\includegraphics[width=1\textwidth]{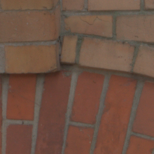}
\end{minipage}
\begin{minipage}{0.089\textwidth}
\includegraphics[width=1\textwidth]{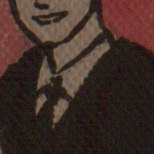}
\end{minipage}
\begin{minipage}{0.089\textwidth}
\includegraphics[width=1\textwidth]{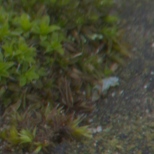}
\end{minipage}
}\vspace{-3mm}
\subfigure{
\begin{minipage}{0.089\textwidth}
\includegraphics[width=1\textwidth]{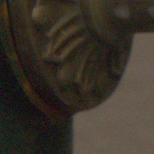}
\end{minipage}
\begin{minipage}{0.089\textwidth}
\includegraphics[width=1\textwidth]{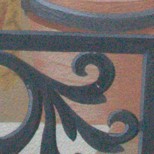}
\end{minipage}
\begin{minipage}{0.089\textwidth}
\includegraphics[width=1\textwidth]{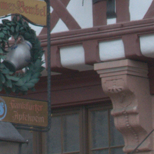}
\end{minipage}
\begin{minipage}{0.089\textwidth}
\includegraphics[width=1\textwidth]{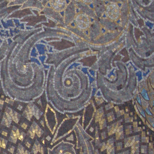}
\end{minipage}
\begin{minipage}{0.089\textwidth}
\includegraphics[width=1\textwidth]{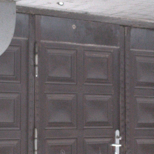}
\end{minipage}
}\vspace{-3mm}
\caption{Some sample images from the \textbf{Dataset 2} \cite{dnd2017}.}
\vspace{-3mm}
\label{fig5}
\vspace{-3mm}
\end{figure}

\textbf{Dataset 2} is called the Darmstadt Noise Dataset (DND) \cite{dnd2017}, which includes 50 different pairs of images of the same scenes captured by Sony A7R, Olympus E-M10, Sony RX100 IV, and Huawei Nexus 6P.\ The real-world noisy images are collected under higher ISO values with shorter exposure time, while the ``ground truth'' images are captured under lower ISO values with longer exposure times. Since the captured images are of megapixel-size, the authors cropped 20 bounding boxes of $512\times512$ pixels from each image in the dataset, yielding $50\times20=1000$ test crops in total. Some samples are shown in Fig.\ \ref{fig5}. Note that the ``ground truth'' images of this dataset have not been released yet, but one can submit the denoised images to the \href{https://noise.visinf.tu-darmstadt.de/}{Project Website} and get the average PSNR (dB) and SSIM results.

\begin{figure}[t]
\vspace{-3mm}
\centering
\subfigure{
\begin{minipage}{0.089\textwidth}
\includegraphics[width=1\textwidth]{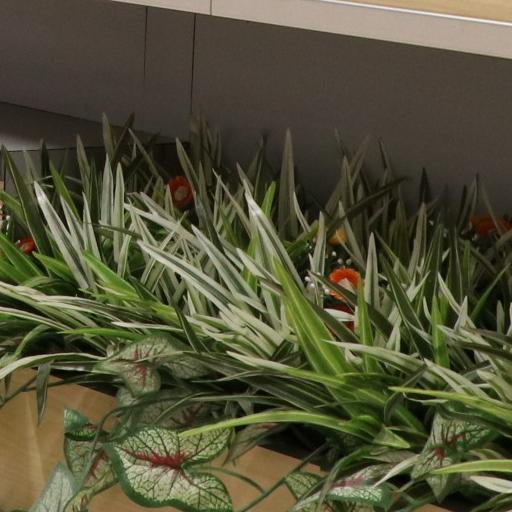}
\end{minipage}
\begin{minipage}{0.089\textwidth}
\includegraphics[width=1\textwidth]{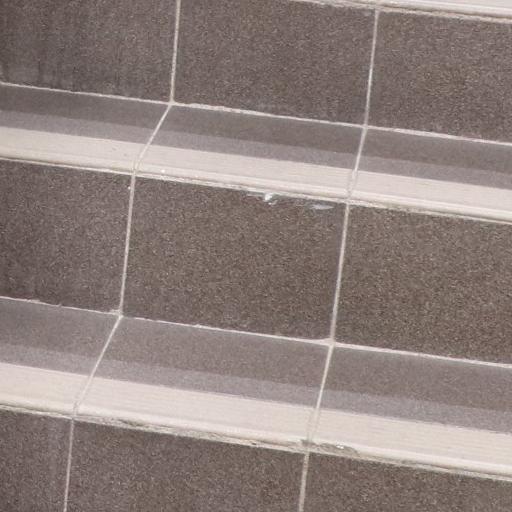}
\end{minipage}
\begin{minipage}{0.089\textwidth}
\includegraphics[width=1\textwidth]{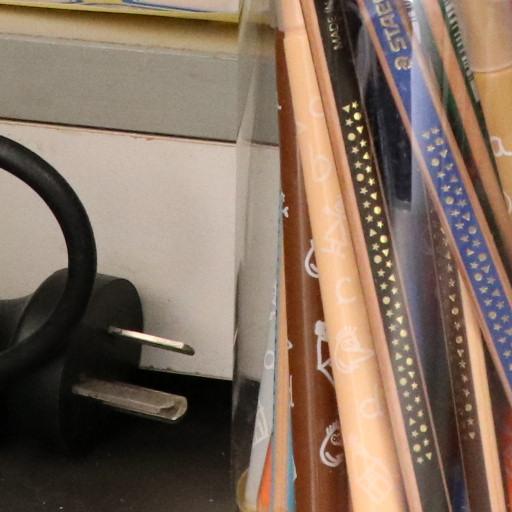}
\end{minipage}
\begin{minipage}{0.089\textwidth}
\includegraphics[width=1\textwidth]{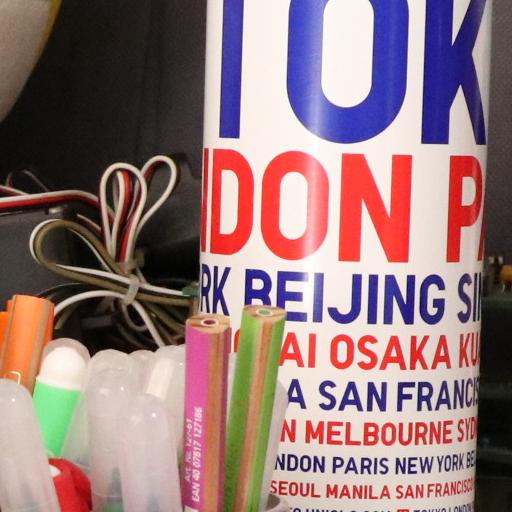}
\end{minipage}
\begin{minipage}{0.089\textwidth}
\includegraphics[width=1\textwidth]{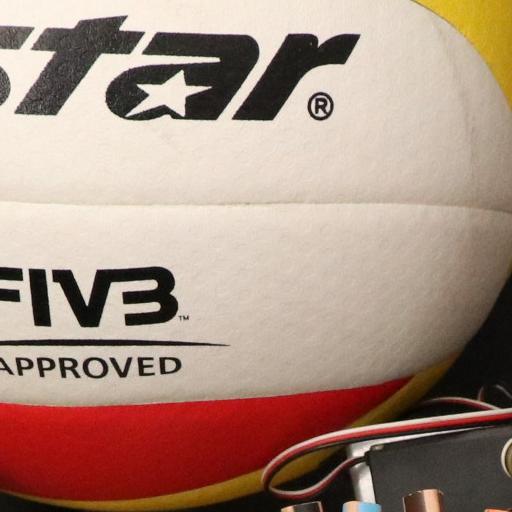}
\end{minipage}
}\vspace{-3mm}
\subfigure{
\begin{minipage}{0.089\textwidth}
\includegraphics[width=1\textwidth]{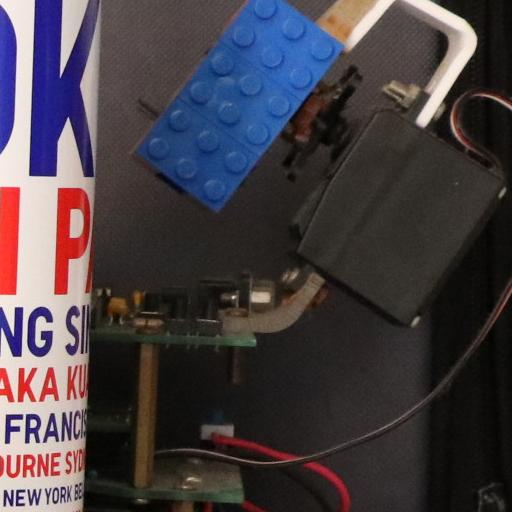}
\end{minipage}
\begin{minipage}{0.089\textwidth}
\includegraphics[width=1\textwidth]{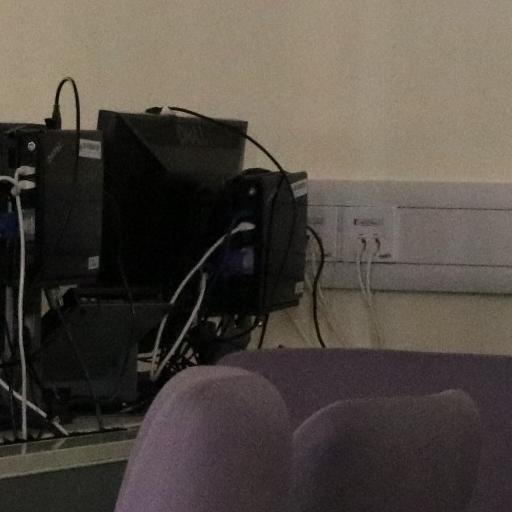}
\end{minipage}
\begin{minipage}{0.089\textwidth}
\includegraphics[width=1\textwidth]{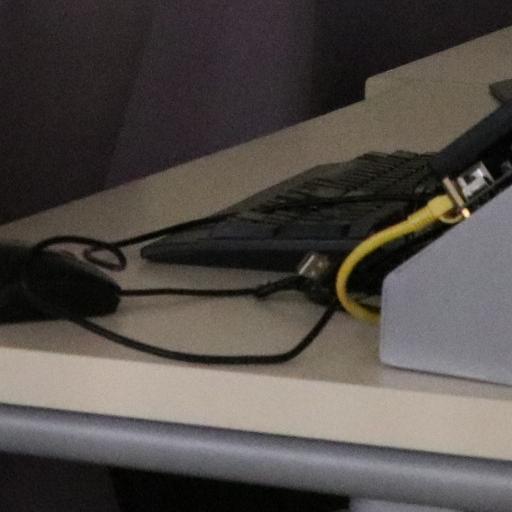}
\end{minipage}
\begin{minipage}{0.089\textwidth}
\includegraphics[width=1\textwidth]{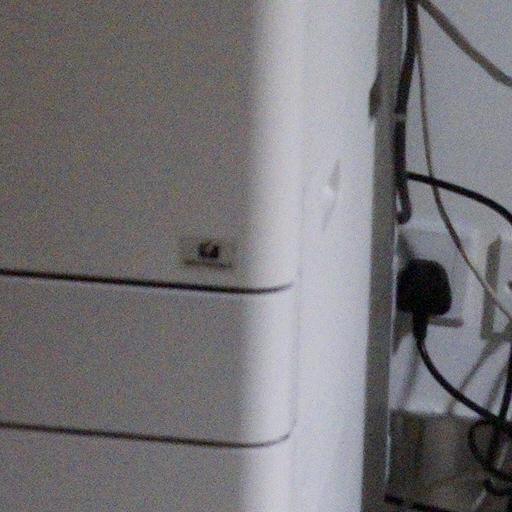}
\end{minipage}
\begin{minipage}{0.089\textwidth}
\includegraphics[width=1\textwidth]{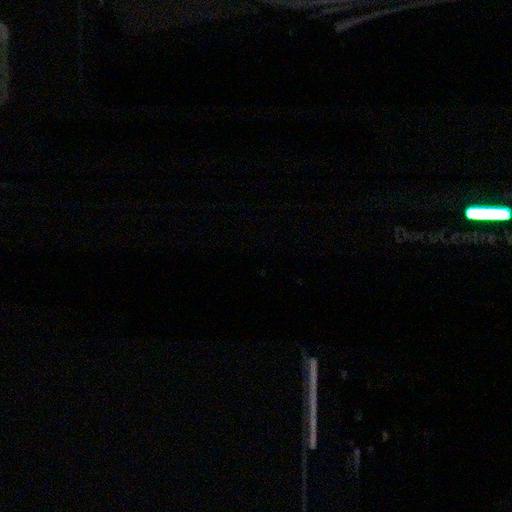}
\end{minipage}
}\vspace{-3mm}
\caption{Some sample images from our dataset (\textbf{Dataset 3}).}
\vspace{-3mm}
\label{fig6}
\vspace{-3mm}
\end{figure}

\textbf{Dataset 3.} On one hand, the scenes of \textbf{Dataset 1} are mostly printed photos, and they cannot represent realistic objects and scenes with different reflectance properties. On the other hand, the \textbf{Dataset 2} contains repetitive contents in the 20 cropped images for each of the 50 scenes. To remedy the limitations of \textbf{Dataset 1} and \textbf{Dataset 2}, we construct another dataset which contains images of 10 different scenes captured by Canon 80D and Sony A7II cameras with more ISO settings and more comprehensive scenes. The ISO settings in our dataset are 800, 1600, 3200, 6400, 12800 while those of \textbf{Dataset 1} are 1600, 3200, 6400. Compared to \textbf{Dataset 2}, our new dataset is more comprehensive on scene contents. Similar to \textbf{Dataset 1}, each scene was captured 500 shots, and the mean image of these 500 shots can be used a kind of ground-truth to evaluate the denoising algorithms. Fig.\ \ref{fig6} shows some cropped images of the scenes in our dataset. One can see that the images contain a lot of different realistic objects with varying colors, shapes, materials, etc.

Our dataset provides real-world noisy images of realistic objects with different ISO settings. It can be used to more fairly evaluate the performance of different real-world noisy image denoising methods. Consider that the image resolution is very high (about $4000\times4000$), for the convenience of experimental studies, we cropped 100 (10 for each scene) smaller images (of size $512\times512$) from it to perform experiments. The whole dataset will be made publically available with the publication of this paper.

\begin{figure*}
\vspace{-4mm}
\centering
\subfigure{
\begin{minipage}[t]{0.19\textwidth}
\centering
\raisebox{-0.15cm}{\includegraphics[width=1\textwidth]{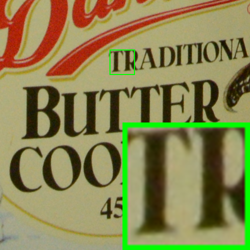}}
{\footnotesize (a) Noisy \cite{crosschannel2016}: 35.89dB  }
\end{minipage}
\begin{minipage}[t]{0.19\textwidth}
\centering
\raisebox{-0.15cm}{\includegraphics[width=1\textwidth]{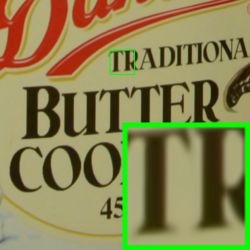}}
{\footnotesize (b) External: 39.05dB }
\end{minipage}
\begin{minipage}[t]{0.19\textwidth}
\centering
\raisebox{-0.15cm}{\includegraphics[width=1\textwidth]{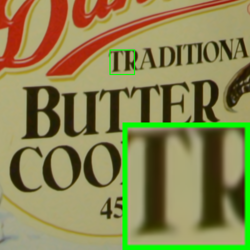}}
{\footnotesize (c) Internal: 38.75dB }
\end{minipage}
\begin{minipage}[t]{0.19\textwidth}
\centering
\raisebox{-0.15cm}{\includegraphics[width=1\textwidth]{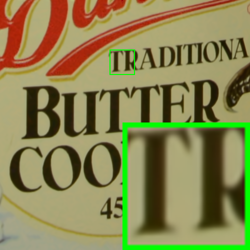}}
{\footnotesize (d) Guided Internal: \textbf{39.39}dB }
\end{minipage}
\begin{minipage}[t]{0.19\textwidth}
\centering
\raisebox{-0.15cm}{\includegraphics[width=1\textwidth]{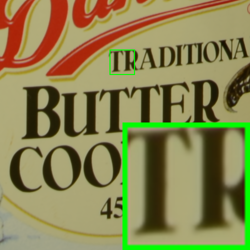}}
{\footnotesize (e) Mean Image \cite{crosschannel2016}}
\end{minipage}
}\vspace{-3mm}
\caption{Denoised images of a region cropped from the real-world noisy image ``Nikon D600 ISO 3200 C1'' \cite{crosschannel2016} by different methods. The images are better to be zoomed-in on screen.}
\label{fig7}
\vspace{-1mm}
\end{figure*}

\begin{figure*}
\centering
\subfigure{
\begin{minipage}[t]{0.19\textwidth}
\centering  
\raisebox{-0.15cm}{\includegraphics[width=1\textwidth]{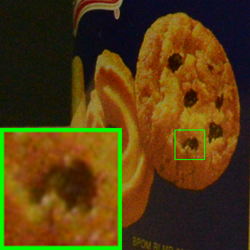}}
{\footnotesize (a) Noisy \cite{crosschannel2016}: 33.77dB  }
\end{minipage}
\begin{minipage}[t]{0.19\textwidth} 
\centering
\raisebox{-0.15cm}{\includegraphics[width=1\textwidth]{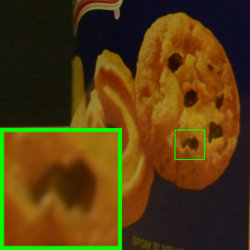}}
{\footnotesize (b) External: 36.97dB }
\end{minipage}
\begin{minipage}[t]{0.19\textwidth}
\centering
\raisebox{-0.15cm}{\includegraphics[width=1\textwidth]{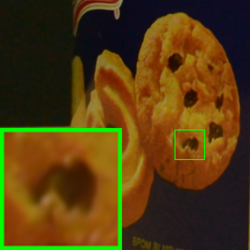}}
{\footnotesize (c) Internal: 37.40dB }
\end{minipage} 
\begin{minipage}[t]{0.19\textwidth}
\centering
\raisebox{-0.15cm}{\includegraphics[width=1\textwidth]{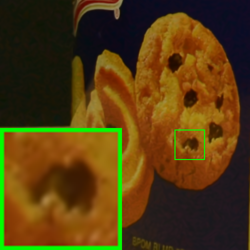}}
{\footnotesize (d) Guided Internal: \textbf{38.01}dB }
\end{minipage}
\begin{minipage}[t]{0.19\textwidth}
\centering
\raisebox{-0.15cm}{\includegraphics[width=1\textwidth]{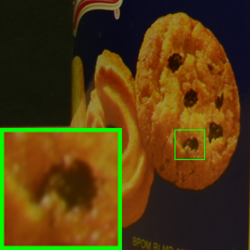}}
{\footnotesize (e) Mean Image \cite{crosschannel2016}}
\end{minipage}
}\vspace{-3mm}
\caption{Denoised images of a region cropped from the real-world noisy image ``Nikon D600 ISO 3200 C1'' \cite{crosschannel2016} by different methods. The images are better to be zoomed-in on screen.}
\label{fig8}
\vspace{-2mm}
\end{figure*}

\subsection{Comparison among external, internal and guided internal priors}

To demonstrate the advantages of external prior guided internal prior learning, we perform real-world noisy image denoising by using external priors only (denoted by ``External''), internal priors only (denoted by ``Internal''), and the proposed guided internal priors (denoted by ``Guided Internal"), respectively.\ For the ``External'' method, we utilize the full external dictionaries (i.e., $r=108$ in Eq.\ (\ref{equ5})) for denoising.\ For the ``Internal'' method, the overall framework is similar to the method of \cite{ncsr}.\ A GMM model (with $K = 32$ Gaussians) is directly learned from the PGs extracted from the given noisy image without using any external data, and then the internal orthogonal dictionaries are obtained via Eq.\ (\ref{equ2}) to perform denoising.\ All parameters of the ``External'' and ``Internal'' methods are tuned to achieve their best performance. 

We compare the three methods on the 60 cropped images from \textbf{Dataset 1} \cite{crosschannel2016}.\ The average PSNR and run time are listed in Table \ref{tab1}.\ The best results are highlighted in bold.\ It can be seen that ``Guided Internal'' method achieves better PSNR than both ``External'' and ``Internal'' methods. In addition, the ``Internal'' method is very slow because it involves online GMM learning, while the ``Guided Internal'' method is only a little slower than the ``External'' method.\ Figs. \ref{fig7} and \ref{fig8} show the denoised images of two noisy images by the three methods.\ One can see that the  ``External'' method is good at recovering large-scale structures (see Fig. \ref{fig7}) while the ``Internal'' method is good at recovering fine-scale textures (see Fig. \ref{fig8}).\ By utilizing external priors to guide the internal prior learning, our proposed method can effectively recover both the large-scale structures and fine-scale textures. 

\begin{table}[ht!]
\vspace{-1mm}
\caption{Average PSNR (dB) and Run Time (seconds) of the ``External'', ``Internal'', and ``Guided Internal'' methods on 60 real-world noisy images (of size $500\times500\times3$) cropped from \textbf{Dataset 1} \cite{crosschannel2016}.}
\vspace{-4mm}
\label{tab1}
\begin{center}
\renewcommand\arraystretch{1}
\begin{tabular}{|c||c|c|c|c|}
\hline
 & \small\textbf{Noisy} &\small \textbf{External} & \small\textbf{Internal} & \small\textbf{Guided Internal}  
\\
\hline
PSNR & 34.51 & 38.21 & 38.07 & \textbf{38.75} 
\\
\hline
Time & | &  \textbf{21.19}  & 312.67 & 22.26
\\
\hline
\end{tabular}
\end{center}
\vspace{-6mm}
\end{table}

\vspace{-1mm}
\subsection{Comparison with State-of-the-Art Denoising Methods}

\textbf{Comparison methods.}
We compare the proposed method with state-of-the-art image denoising methods, including GAT-BM3D \cite{Makitalo2013Optimal}, CBM3D \cite{cbm3d}, WNNM \cite{wnnm}, TID \cite{tid}, MLP \cite{mlp}, DnCNN \cite{dncnn}, CSF \cite{csf}, TNRD \cite{chen2015learning}, Noise Clinic (NC) \cite{noiseclinic,ncwebsite}, Cross-Channel (CC) \cite{crosschannel2016}, and Neat Image (NI) \cite{neatimage}.\ Among these methods, GAT-BM3D \cite{Makitalo2013Optimal} is a state-of-the-art Poisson noise reduction method.\ The method CBM3D \cite{cbm3d} is a state-of-the-art method for color image denoising and the noise on color images is assumed to be additive white Gaussian.\ The methods of WNNM, MLP, DnCNN, CSF, and TNRD are state-of-the-art Gaussian noise removal methods for grayscale images, and we apply them to each channel of color images for denoising. NC is a blind image denoising method, and NI is a set of commercial software for image denoising, which has been embedded into Photoshop and Corel PaintShop. The code of CC is not released but its results on the 15 cropped images are available at \cite{crosschannel2016}. Therefore, we only compare with it on the 15 cropped images in \textbf{Dataset 1} \cite{crosschannel2016}. 

\textbf{Noise level of comparison methods.} 
For the CBM3D method, the standard deviation of noise on color images should be given as a parameter.\ For methods of WNNM, MLP, CSF, and TNRD, the noise level in each color channel should be input. For the DnCNN method, it is trained to deal with noise in a range of levels $0\sim55$.\ We retrain the models of discriminative denoising methods MLP, CSF, and TNRD (using the released codes by the authors) at different noise levels from $\sigma=5$ to $\sigma=50$ with a gap of $5$.\ The denoising is performed by processing each channel with the model trained at the same (or nearest) noise level.\ The noise levels ($\sigma_{r}, \sigma_{g}, \sigma_{b}$) in R, G ,B channels are assumed to be Gaussian and can be estimated via some noise estimation methods \cite{noiselevel,Chen2015ICCV}.\ In this paper, we employ the method \cite{Chen2015ICCV} to estimate the noise level for each color channel.

\begin{table*}
\vspace{-0mm}
\small
\caption{PSNR(dB) results and Speed (sec.) of different methods on 15 cropped real-world noisy images used in \cite{crosschannel2016}.}
\vspace{-4mm}
\label{tab2}
\begin{center}
\renewcommand\arraystretch{1}
\begin{tabular}{|c||c|c|c|c|c|c|c|c|c|c|c|c|}
\hline 
Setting
&\textbf{GAT-BM3D}
&\textbf{CBM3D}
&\textbf{WNNM}
&\textbf{TID}
&\textbf{MLP}
&\textbf{CSF}
&\textbf{TNRD}
&\textbf{DnCNN}
& \textbf{NI}
&\textbf{NC}
&\textbf{CC} 
&\textbf{Ours} 
\\
\hline
\multirow{3}{*}{\small{Canon 5D}} 
& 31.23 & 39.76 & 37.51 & 37.22 & 39.00 & 35.68 & 39.51 & 37.26 & 37.68 & 38.76 & 38.37 & \textbf{40.50}
\\ 
\cdashline{2-13} 
\multirow{3}{*}{ISO = 3200}   
& 30.55 & 36.40 & 33.86 & 34.54 & 36.34 & 34.03 & 36.47 & 34.13 & 34.87 & 35.69 & 35.37 & \textbf{37.05}
\\ 
\cdashline{2-13}    
& 27.74 & 36.37 & 31.43 & 34.25 & 36.33 & 32.63 & \textbf{36.45} & 34.09 & 34.77 & 35.54 & 34.91 & 36.11 
\\
\hline
\multirow{3}{*}{Nikon D600} 
& 28.55 & 34.18 & 33.46 & 32.99 & 34.70 & 31.78 & 34.79 & 33.62 & 34.12 & \textbf{35.57} & 34.98 & 34.88
\\ 
\cdashline{2-13} 
\multirow{3}{*}{ISO = 3200}   
& 32.01 & 35.07 & 36.09 & 34.20 & 36.20 & 35.16 & 36.37 & 34.48 & 35.36 & \textbf{36.70} & 35.95 & 36.31
\\ 
\cdashline{2-13}    
& 39.78 & 37.13 & 39.86 & 35.58 & 39.33 & 39.98 & 39.49 & 35.41 & 38.68 & 39.28 & \textbf{41.15} & 39.23
\\
\hline
\multirow{3}{*}{Nikon D800} 
& 32.24 & 36.81 & 36.35 & 34.94 & 37.95 & 34.84 & 38.11 & 35.79 & 37.34 & 38.01 & 37.99 & \textbf{38.40}
\\ 
\cdashline{2-13} 
\multirow{3}{*}{ISO = 1600}   
& 33.86 & 37.76 & 39.99 & 35.19 & 40.23 & 38.42 & 40.52 & 36.08 & 38.57 & 39.05 & 40.36 & \textbf{40.92}
\\ 
\cdashline{2-13}    
& 33.90 & 37.51 & 37.15 & 35.26 & 37.94 & 35.79 & 38.17 & 35.48 & 37.87 & 38.20 & 38.30 & \textbf{38.97}
\\
\hline
\multirow{3}{*}{Nikon D800} 
& 36.49 & 35.05 & 38.60 & 33.70 & 37.55 & 38.36 & 37.69 & 34.08 & 36.95 & 38.07 & \textbf{39.01} & 38.66
\\ 
\cdashline{2-13} 
\multirow{3}{*}{ISO = 3200}   
& 32.91 & 34.07 & 36.04 & 31.04 & 35.91 & 35.53 & 35.90 & 33.70 & 35.09 & 35.72 & 36.75 & \textbf{37.07}
\\ 
\cdashline{2-13}    
& \textbf{40.20} & 34.42 & 39.73 & 33.07 & 38.15 & 40.05 & 38.21 & 33.31 & 36.91 & 36.76 & 39.06 & 38.52
\\ 
\hline
\multirow{3}{*}{Nikon D800} 
& 29.84 & 31.13 & 33.29 & 29.40 & 32.69 & 34.08 & 32.81 & 29.83 & 31.28 & 33.49 & \textbf{34.61} & 33.76
\\ 
\cdashline{2-13} 
\multirow{3}{*}{ISO = 6400}   
& 27.94 & 31.22 & 31.16 & 29.86 & 32.33 & 32.13 & 32.33 & 30.55 & 31.38 & 32.79 & 33.21 & \textbf{33.43}
\\ 
\cdashline{2-13}    
& 29.15 & 30.97 & 31.98 & 29.21 & 32.29 & 31.52 & 32.29 & 30.09 & 31.40 & 32.86 & 33.22 & \textbf{33.58}
\\
\hline
Average
& 32.43 & 35.19 & 35.77 & 33.36 & 36.46 & 35.33 & 36.61 & 33.86 & 35.49 & 36.43 & 36.88 & \textbf{37.15}
\\
\hline
Time (s)
& 10.9 & 6.9 & 151.5 & 7353.2 & 16.8 & 19.3 & 5.1 & 79.2 & \textbf{0.6} & 15.3 & NA & 23.9
\\
\hline
\end{tabular}
\end{center}
\vspace{-3mm}
\end{table*}

\textbf{Results on Dataset 1.}
As described in section 4.2, there is a mean image for each of the 11 scenes used in \textbf{Dataset 1} \cite{crosschannel2016}, and those mean images can be roughly taken as ``ground truth" images for quantitative evaluation of denoising algorithms. We firstly perform quantitative comparison on the 15 cropped images used in \cite{crosschannel2016}. The results on PSNR (dB) and speed (second) of GAT-BM3D, CBM3D, WNNM, TID, MLP, CSF, TNRD, DnCNN, NC, NI and CC are listed in Table II (The results of CC are copied from the original paper \cite{crosschannel2016}). The best PSNR results of each image are highlighted in bold. One can see that on 8 out of the 15 images, our method achieves the best PSNR values. CC achieves the best PSNR on 3 of the 15 images. It should be noted that in the CC method, a specific model is trained for each camera and camera setting, while our method uses the same model for all images. On average, our proposed method has 0.27dB PSNR improvements over the second best method CC and much higher PSNR gains over other competing methods. The method GAT-BM3D does not work well on most images. This is because real world noise is much more complex than Poisson. 

\begin{figure*}
\vspace{-1mm}
\centering
\subfigure{
\begin{minipage}[t]{0.19\textwidth}
\centering
\raisebox{-0.15cm}{\includegraphics[width=1\textwidth]{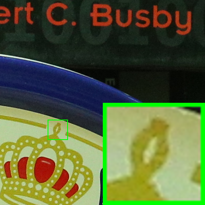}}
\centering{\footnotesize (a) Noisy  \cite{crosschannel2016}: 37.00dB }
\end{minipage}
\begin{minipage}[t]{0.19\textwidth}
\centering
\raisebox{-0.15cm}{\includegraphics[width=1\textwidth]{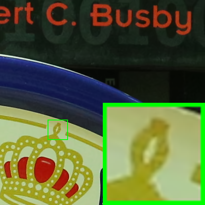}}
\centering{\footnotesize (b) CBM3D \cite{cbm3d}: 39.76dB}
\end{minipage}
\begin{minipage}[t]{0.19\textwidth}
\centering
\raisebox{-0.15cm}{\includegraphics[width=1\textwidth]{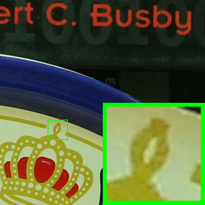}}
\centering{\footnotesize (c) TID \cite{tid}: 37.22dB}
\end{minipage}
\centering
\begin{minipage}[t]{0.19\textwidth}
\raisebox{-0.15cm}{\includegraphics[width=1\textwidth]{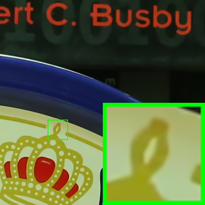}}
\centering{\footnotesize (d) TNRD \cite{chen2015learning}: 39.51dB  } 
\end{minipage}
\begin{minipage}[t]{0.19\textwidth}
\centering
\raisebox{-0.15cm}{\includegraphics[width=1\textwidth]{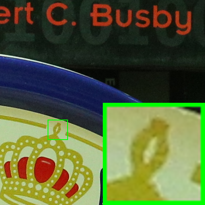}}
\centering{\footnotesize (e) DnCNN \cite{dncnn}: 37.26dB }
\end{minipage}
}\vspace{-3mm}
\subfigure{
\begin{minipage}[t]{0.19\textwidth}
\centering
\raisebox{-0.15cm}{\includegraphics[width=1\textwidth]{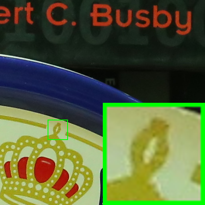}}
\centering{\footnotesize (f) NI \cite{neatimage}: 37.68dB  }
\end{minipage}
\begin{minipage}[t]{0.19\textwidth}
\centering
\raisebox{-0.15cm}{\includegraphics[width=1\textwidth]{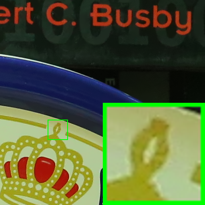}}
\centering{\footnotesize (g) NC \cite{noiseclinic,ncwebsite}: 38.76dB  }
\end{minipage}
\begin{minipage}[t]{0.19\textwidth}
\centering
\raisebox{-0.15cm}{\includegraphics[width=1\textwidth]{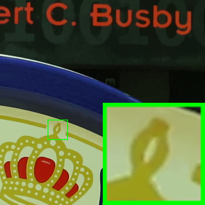}}
\centering{\footnotesize (h) CC \cite{crosschannel2016}: 38.37dB }
\end{minipage}
\begin{minipage}[t]{0.19\textwidth}
\centering
\raisebox{-0.15cm}{\includegraphics[width=1\textwidth]{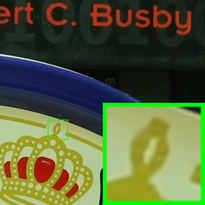}}
\centering{\footnotesize (i) Ours: \textbf{40.50}dB}
\end{minipage}
\begin{minipage}[t]{0.19\textwidth}
\centering
\raisebox{-0.15cm}{\includegraphics[width=1\textwidth]{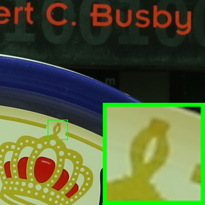}}
\centering{\footnotesize (j) Mean Image \cite{crosschannel2016}}
\end{minipage}
}\vspace{-3mm}
\caption{Denoised images of a region cropped from the real-world noisy image ``Canon 5D Mark 3 ISO 3200 1" \cite{crosschannel2016} by different methods. The images are better to be zoomed-in on screen.}
\label{fig9}
\vspace{-3mm}
\end{figure*}

\begin{figure*}
\centering
\subfigure{
\begin{minipage}[t]{0.19\textwidth}
\centering
\raisebox{-0.15cm}{\includegraphics[width=1\textwidth]{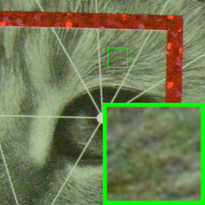}}
\centering{\footnotesize (a) Noisy  \cite{crosschannel2016}: 37.00dB }
\end{minipage}
\begin{minipage}[t]{0.19\textwidth}
\centering
\raisebox{-0.15cm}{\includegraphics[width=1\textwidth]{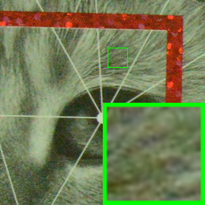}}
\centering{\footnotesize (b) CBM3D \cite{cbm3d}: 39.76dB}
\end{minipage}
\begin{minipage}[t]{0.19\textwidth}
\centering
\raisebox{-0.15cm}{\includegraphics[width=1\textwidth]{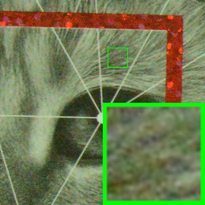}}
\centering{\footnotesize (c) TID \cite{tid}: 37.51dB}
\end{minipage}
\centering
\begin{minipage}[t]{0.19\textwidth}
\raisebox{-0.15cm}{\includegraphics[width=1\textwidth]{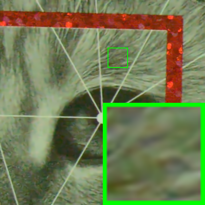}}
\centering{\footnotesize (d) TNRD \cite{chen2015learning}: 39.51dB  } 
\end{minipage}
\begin{minipage}[t]{0.19\textwidth}
\centering
\raisebox{-0.15cm}{\includegraphics[width=1\textwidth]{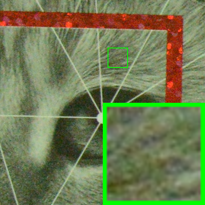}}
\centering{\footnotesize (e) DnCNN \cite{dncnn}: 37.26dB }
\end{minipage}
}\vspace{-3mm}
\subfigure{
\begin{minipage}[t]{0.19\textwidth}
\centering
\raisebox{-0.15cm}{\includegraphics[width=1\textwidth]{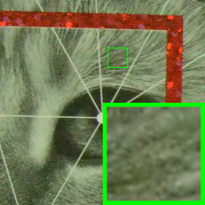}}
\centering{\footnotesize (f) NI \cite{neatimage}: 37.68dB  }
\end{minipage}
\begin{minipage}[t]{0.19\textwidth}
\centering
\raisebox{-0.15cm}{\includegraphics[width=1\textwidth]{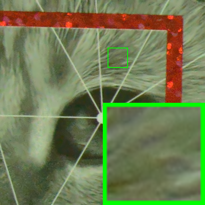}}
\centering{\footnotesize (g) NC \cite{noiseclinic,ncwebsite}: 38.76dB  }
\end{minipage}
\begin{minipage}[t]{0.19\textwidth}
\centering
\raisebox{-0.15cm}{\includegraphics[width=1\textwidth]{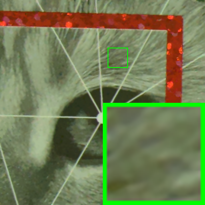}}
\centering{\footnotesize (h) CC \cite{crosschannel2016}: 38.37dB }
\end{minipage}
\begin{minipage}[t]{0.19\textwidth}
\centering
\raisebox{-0.15cm}{\includegraphics[width=1\textwidth]{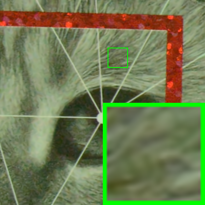}}
\centering{\footnotesize (i) Ours: \textbf{40.50}dB}
\end{minipage}
\begin{minipage}[t]{0.19\textwidth}
\centering
\raisebox{-0.15cm}{\includegraphics[width=1\textwidth]{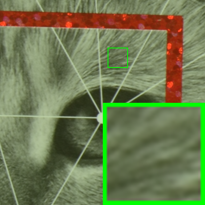}}
\centering{\footnotesize (j) Mean Image \cite{crosschannel2016}}
\end{minipage}
}\vspace{-3mm}
\caption{Denoised images of a region cropped from the real-world noisy image ``Nikon D800 ISO 6400 1'' \cite{crosschannel2016} by different methods. The images are better to be zoomed-in on screen.}
\label{fig10}
\vspace{-4mm}
\end{figure*}

Figs. \ref{fig9} and \ref{fig10} show the denoised images of one scene captured by Canon 5D Mark 3 at ISO = 3200 and Nikon D800 at ISO = 6400, respectively. We can see that GAT-BM3D, CBM3D, TID, DnCNN, NC, NI and CC would either remain noise or generate artifacts, while TNRD over-smooths much the image. By using the external prior guided internal priors, our proposed method preserves edges and textures better than other methods while removing the noise, leading to visually more pleasant outputs.\ Specifically, Fig.\ \ref{fig10} is used to illustrate the denoising performance of our method on fine-scale textures such as hair, which is a very challenging task.\ Even the ``ground truth'' mean image cannot show very clear details of the hair.\ Though our method cannot reproduce clearly the details (e.g., the local direction of hair in some regions), it demonstrates the best visual results among the competing methods.\ More comparisons on visual quality and SSIM \cite{ssim} index can be found in the supplementary file.

\begin{figure*}
\vspace{-4mm}
\centering
\subfigure{
\begin{minipage}[t]{0.196\textwidth}
\centering
\raisebox{-0.15cm}{\includegraphics[width=1\textwidth]{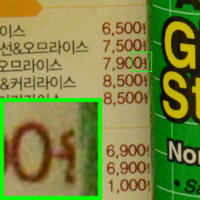}}
{\footnotesize (a) Noisy \cite{crosschannel2016}: 33.60dB }
\end{minipage}
\begin{minipage}[t]{0.196\textwidth}
\centering
\raisebox{-0.15cm}{\includegraphics[width=1\textwidth]{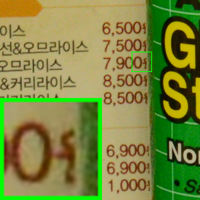}}
{\footnotesize (b) CBM3D \cite{cbm3d}: 35.23dB  }
\end{minipage}
\begin{minipage}[t]{0.196\textwidth}
\centering
\raisebox{-0.15cm}{\includegraphics[width=1\textwidth]{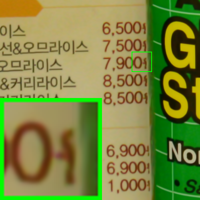}}
{\footnotesize (c) WNNM \cite{wnnm}: 36.50dB  }
\end{minipage}
\begin{minipage}[t]{0.196\textwidth}
\centering
\raisebox{-0.15cm}{\includegraphics[width=1\textwidth]{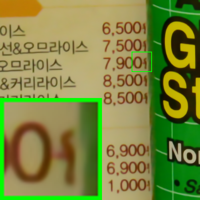}}
{\footnotesize (d) CSF \cite{csf}: 36.21dB }
\end{minipage}
\begin{minipage}[t]{0.196\textwidth}
\centering
\raisebox{-0.15cm}{\includegraphics[width=1\textwidth]{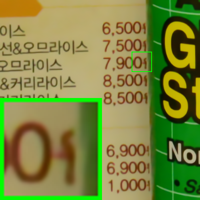}}
{\footnotesize (e) TNRD \cite{chen2015learning}: 37.10dB   }
\end{minipage}
}\vspace{-3mm}
\subfigure{
\begin{minipage}[t]{0.196\textwidth}
\centering
\raisebox{-0.15cm}{\includegraphics[width=1\textwidth]{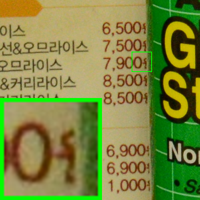}}
{\footnotesize (f) DnCNN \cite{dncnn}: 34.43dB }
\end{minipage}
\begin{minipage}[t]{0.196\textwidth}
\centering
\raisebox{-0.15cm}{\includegraphics[width=1\textwidth]{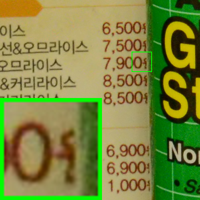}}
{\footnotesize (g) NI \cite{neatimage}: 35.02dB  }
\end{minipage}
\begin{minipage}[t]{0.196\textwidth}
\centering
\raisebox{-0.15cm}{\includegraphics[width=1\textwidth]{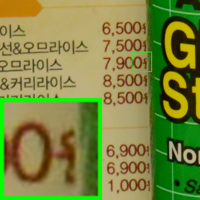}}
{\footnotesize (h) NC \cite{ncwebsite,noiseclinic}: 36.07dB   }
\end{minipage}
\begin{minipage}[t]{0.196\textwidth}
\centering
\raisebox{-0.15cm}{\includegraphics[width=1\textwidth]{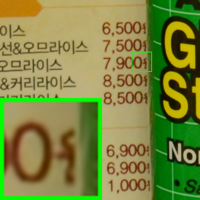}}
{\footnotesize (i) Ours: \textbf{37.50}dB  }
\end{minipage}
\begin{minipage}[t]{0.196\textwidth}
\centering
\raisebox{-0.15cm}{\includegraphics[width=1\textwidth]{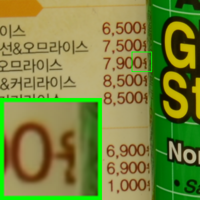}}
{\footnotesize (j) Mean Image \cite{crosschannel2016} }
\end{minipage}
}\vspace{-3mm}
\caption{Denoised images of a region cropped from the real-world noisy image ``Nikon D800 ISO 3200 A3" \cite{crosschannel2016} by different methods. The images are better viewed by zooming in on screen.} 
\vspace{-3mm}
\label{fig11}
\end{figure*}

We then perform denoising experiments on the 60 images we cropped from \cite{crosschannel2016}. The average PSNR results are listed in Table III (CC is not compared since the code is not available). Again, our proposed method achieves much better PSNR results than the other methods. The improvements of our method over the second best method (TNRD) are 0.43dB on PSNR. Fig. \ref{fig11} shows the denoised images of one scene captured by Nikon D800 at ISO = 3200. We can see again that the proposed method obtain better visual quality than other competing methods. More comparisons on visual quality and SSIM can be found in the supplementary file.

\begin{figure*}
\vspace{-0mm}
\centering
\subfigure{
\begin{minipage}[t]{0.19\textwidth}
\centering
\raisebox{-0.15cm}{\includegraphics[width=1\textwidth]{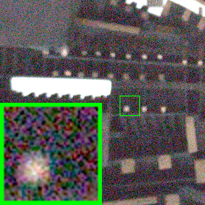}}
{\footnotesize (a) Noisy \cite{ncwebsite}   }
\end{minipage}
\begin{minipage}[t]{0.19\textwidth}
\centering
\raisebox{-0.15cm}{\includegraphics[width=1\textwidth]{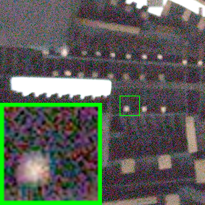}}
{\footnotesize (b) CBM3D \cite{cbm3d}}
\end{minipage}
\begin{minipage}[t]{0.19\textwidth}
\centering
\raisebox{-0.15cm}{\includegraphics[width=1\textwidth]{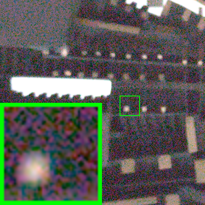}}
{\footnotesize (c) WNNM \cite{wnnm} }
\end{minipage}
\begin{minipage}[t]{0.19\textwidth}
\centering
\raisebox{-0.15cm}{\includegraphics[width=1\textwidth]{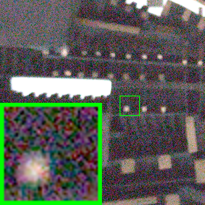}}
{\footnotesize (d) MLP \cite{mlp}   }
\end{minipage}
\begin{minipage}[t]{0.19\textwidth}
\centering
\raisebox{-0.15cm}{\includegraphics[width=1\textwidth]{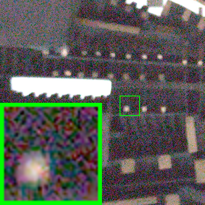}}
{\footnotesize (e) CSF \cite{csf}}
\end{minipage}
}\vspace{-3mm}
\subfigure{
\begin{minipage}[t]{0.19\textwidth}
\centering
\raisebox{-0.15cm}{\includegraphics[width=1\textwidth]{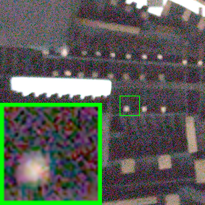}}
{\footnotesize (f) TNRD \cite{chen2015learning}}
\end{minipage}
\begin{minipage}[t]{0.19\textwidth}
\centering
\raisebox{-0.15cm}{\includegraphics[width=1\textwidth]{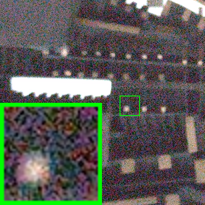}}
{\footnotesize (g) DnCNN \cite{dncnn}  }
\end{minipage}
\begin{minipage}[t]{0.19\textwidth}
\centering
\raisebox{-0.15cm}{\includegraphics[width=1\textwidth]{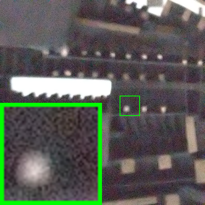}}
{\footnotesize (h) NI \cite{neatimage}  }
\end{minipage}
\begin{minipage}[t]{0.19\textwidth}
\centering
\raisebox{-0.15cm}{\includegraphics[width=1\textwidth]{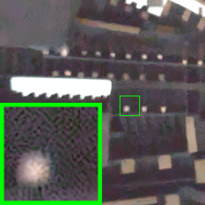}}
{\footnotesize (i) NC \cite{noiseclinic,ncwebsite}   }
\end{minipage}
\begin{minipage}[t]{0.19\textwidth}
\centering
\raisebox{-0.15cm}{\includegraphics[width=1\textwidth]{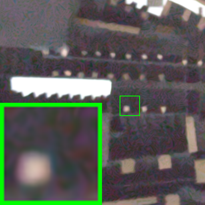}}
{\footnotesize (j) Ours  }
\end{minipage}
}\vspace{-3mm}
\caption{Denoised images by different methods of the real-world noisy image ``0001\_2'' captured by a Huawei Nexus 6P phone\cite{dnd2017}. Note that the ground-truth clean image of the noisy input is not publicly released yet.}
\label{fig12}
\vspace{-2mm}
\end{figure*}

\textbf{Results on Dataset 2.}
In Table \ref{tab4}, we list the average PSNR (dB) results of the competing methods on the 1000 cropped images in the DND dataset \cite{dnd2017}.\ We can see again that the proposed method achieves better performance than the other competing methods.\ Note that the ``ground truth'' images of this dataset have not been released yet, so we are not able to calculate the PSNR and SSIM results for each noisy image in this dataset, nor compare with the ``ground truth'' mean image. However, one can submit the denoised images to the project website and get the average PSNR and SSIM results on the whole 1000 images.\ Fig.\ \ref{fig12} shows the denoised images of a scene ``0001\_2'' captured by a Nexus 6P phone \cite{dnd2017}.\ The noise level in this image is relatively high.\ Hence, this image can be used to justify the performance of the proposed method on real-world noisy images with lower PSNR (around 20dB).\ One can see that the proposed method achieves visually more pleasing results than the other denoising methods.\ More comparisons on visual quality and SSIM can be found in the supplementary file.

\begin{figure*}[ht!]
\centering
\subfigure{
\begin{minipage}[t]{0.196\textwidth}
\centering
\raisebox{-0.15cm}{\includegraphics[width=1\textwidth]{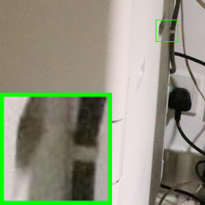}}
{\footnotesize (a) Noisy \cite{crosschannel2016}: 36.51dB }
\end{minipage}
\begin{minipage}[t]{0.196\textwidth}
\centering
\raisebox{-0.15cm}{\includegraphics[width=1\textwidth]{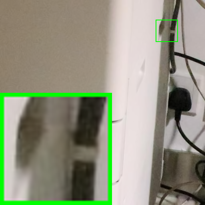}}
{\footnotesize (b) CBM3D \cite{cbm3d}: 37.91dB  }
\end{minipage}
\begin{minipage}[t]{0.196\textwidth}
\centering
\raisebox{-0.15cm}{\includegraphics[width=1\textwidth]{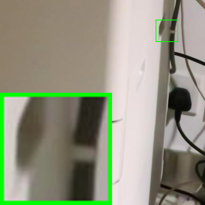}}
{\footnotesize (c) WNNM \cite{wnnm}: 38.23dB  }
\end{minipage}
\begin{minipage}[t]{0.196\textwidth}
\centering
\raisebox{-0.15cm}{\includegraphics[width=1\textwidth]{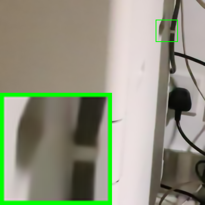}}
{\footnotesize (d) CSF \cite{csf}: 39.02dB }
\end{minipage}
\begin{minipage}[t]{0.196\textwidth}
\centering
\raisebox{-0.15cm}{\includegraphics[width=1\textwidth]{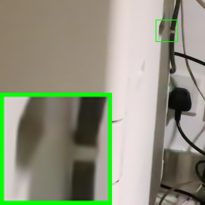}}
{\footnotesize (e) TNRD \cite{chen2015learning}: 39.26dB   }
\end{minipage}
}\vspace{-3mm}
\subfigure{
\begin{minipage}[t]{0.196\textwidth}
\centering
\raisebox{-0.15cm}{\includegraphics[width=1\textwidth]{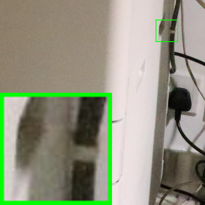}}
{\footnotesize (f) DnCNN \cite{dncnn}: 36.52dB }
\end{minipage}
\begin{minipage}[t]{0.196\textwidth}
\centering
\raisebox{-0.15cm}{\includegraphics[width=1\textwidth]{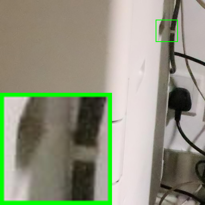}}
{\footnotesize (g) NI \cite{neatimage}: 37.52dB  }
\end{minipage}
\begin{minipage}[t]{0.196\textwidth}
\centering
\raisebox{-0.15cm}{\includegraphics[width=1\textwidth]{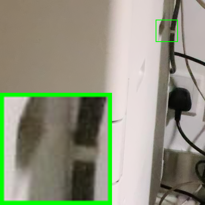}}
\centering{\footnotesize (h) NC \cite{ncwebsite,noiseclinic}: 37.53dB   }
\end{minipage}
\begin{minipage}[t]{0.196\textwidth}
\centering
\raisebox{-0.15cm}{\includegraphics[width=1\textwidth]{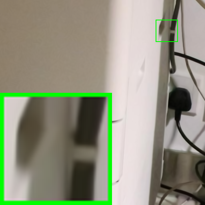}}
{\footnotesize (i) Ours: \textbf{39.41}dB  }
\end{minipage}
\begin{minipage}[t]{0.196\textwidth}
\centering
\raisebox{-0.15cm}{\includegraphics[width=1\textwidth]{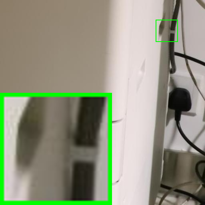}}
{\footnotesize (j) Mean Image }
\end{minipage}
}\vspace{-3mm}
\caption{Denoised images of a region cropped from the real-world noisy image ``Canon 80D ISO 12800 IMG 2321" in our new dataset by different methods. The images are better viewed by zooming in on screen.} 
\label{fig13}
\vspace{-4mm}
\end{figure*}

\textbf{Results on Dataset 3.}
Similar to \textbf{Dataset 1} \cite{crosschannel2016}, there is a ``ground truth'' image for each of the 10 scenes used in our constructed \textbf{Dataset 3}. We perform quantitative comparison on the 100 cropped images. The average PSNR results of competing methods are listed in Table IV. We can see that our proposed method achieves much better PSNR results than the other methods. The improvements of our method over the second best method (TNRD) is 0.16dB on PSNR. Fig. \ref{fig13} shows the denoised images of one scene captured by Canon 80D at ISO = 12800. We can see again that the proposed method removes the noise while maintains better details (such as the vertical black shadow area) than other competing methods. More comparisons on visual quality and SSIM can be found in the supplementary file.

\textbf{Comparison on speed.}
Efficiency is an important aspect to evaluate the efficiency of algorithms. We compare the speed of all competing methods except for CC. All experiments are run under the Matlab2014b environment on a machine with Intel(R) Core(TM) i7-5930K CPU of 3.5GHz and 32GB RAM. The average running time (second) of the compared methods on the 100 real-world noisy images is shown in Table V. The least average running time are highlighted in bold. One can easily see that the commercial software Neat Image (NI) is the fastest method with highly optimized code. For a $512\times512$ image, NI costs about 0.6 second. The other methods cost from 5.2 (TNRD) to 152.2 (WNNM) seconds, while the proposed method costs about 24.1 seconds. It should be noted that GAT-BM3D, CBM3D, TNRD, and NC are implemented with compiled C++ mex-function and with parallelization, while WNNM, TID, MLP, CSF, DnCNN, and the proposed method are implemented purely in Matlab.

\begin{table*}
\vspace{-0mm}
\small
\caption{Average PSNR(dB) results of different methods on 60 real-world noisy images cropped from \cite{crosschannel2016}.}
\vspace{-4mm}
\label{tab3}
\begin{center}
\renewcommand\arraystretch{1}
\begin{tabular}{|c||c|c|c|c|c|c|c|c|c|c|}
\hline
Methods
&\textbf{GAT-BM3D}
&\textbf{CBM3D}
&\textbf{WNNM}
&\textbf{MLP}
&\textbf{CSF} 
&\textbf{TNRD} 
&\textbf{DnCNN}
&\textbf{NI} 
&\textbf{NC} 
&\textbf{Ours} 
\\
\hline
PSNR  
& 34.33 & 36.34 & 37.67 & 38.13 & 37.40 & 38.32 & 34.99 & 36.53 & 37.57 & \textbf{38.75}
\\
\hline
\end{tabular}
\end{center}
\vspace{-3mm}
\end{table*}

\begin{table*}
\vspace{-1mm}
\small
\caption{Average PSNR(dB) results of different methods on the 1000 real-world noisy images from the DND dataset \cite{dnd2017}.}
\vspace{-4mm}
\label{tab4}
\begin{center}
\renewcommand\arraystretch{1}
\begin{tabular}{|c||c|c|c|c|c|c|c|c|c|c|}
\hline
Methods
&\textbf{GAT-BM3D}
&\textbf{CBM3D}
&\textbf{WNNM}
&\textbf{MLP}
&\textbf{CSF} 
&\textbf{TNRD} 
&\textbf{DnCNN}
&\textbf{NI} 
&\textbf{NC} 
&\textbf{Ours} 
\\
\hline
PSNR  
& 30.07 & 32.14 & 33.28 & 34.02 & 33.87 & 34.15 & 32.41 & 35.11 & 36.07 & \textbf{36.41}
\\
\hline
\end{tabular}
\end{center}
\vspace{-3mm}
\end{table*}

\begin{table*}
\vspace{-1mm}
\small
\caption{Average PSNR(dB) results of different methods on 100 real-world noisy images cropped from our new dataset.}
\vspace{-4mm}
\label{tab5}
\begin{center}
\renewcommand\arraystretch{1}
\begin{tabular}{|c||c|c|c|c|c|c|c|c|c|c|}
\hline
Methods
&\textbf{GAT-BM3D}
&\textbf{CBM3D}
&\textbf{WNNM}
&\textbf{MLP}
&\textbf{CSF} 
&\textbf{TNRD} 
&\textbf{DnCNN}
&\textbf{NI} 
&\textbf{NC} 
&\textbf{Ours} 
\\
\hline
PSNR  
& 33.54 & 37.14 & 35.18 & 37.34 & 37.07 & 37.48 & 34.74 & 35.70 & 36.76 & \textbf{37.64}
\\
\hline
\end{tabular}
\end{center}
\vspace{-3mm}
\end{table*}

\begin{table*}
\vspace{-1mm}
\small
\caption{Average Speed (sec.) results of different methods on 100 real-world noisy images cropped from our new dataset.}
\vspace{-4mm}
\label{tab6}
\begin{center}
\renewcommand\arraystretch{1}
\begin{tabular}{|c||c|c|c|c|c|c|c|c|c|c|}
\hline
Methods
&\textbf{GAT-BM3D}
&\textbf{CBM3D}
&\textbf{WNNM}
&\textbf{MLP}
&\textbf{CSF} 
&\textbf{TNRD} 
&\textbf{DnCNN}
&\textbf{NI} 
&\textbf{NC} 
&\textbf{Ours} 
\\
\hline
Time  
& 11.1 & 6.9 & 152.2 & 17.1 & 19.5 & 5.2 & 79.5 & \textbf{0.6} & 15.6 & 24.1
\\
\hline
\end{tabular}
\end{center}
\vspace{-2mm}
\end{table*}

\vspace{-2mm}
\section{Conclusion}
\vspace{-1mm}

We proposed a new prior learning method for the real-world noisy image denoising problem by exploiting the useful information in both external and internal data. We first learned Gaussian Mixture Models (GMMs) from a set of clean external images as general image prior, and then employed the learned GMM model to guide the learning of adaptive internal prior from the given noisy image. Finally, a set of orthogonal dictionaries were output as the external-internal hybrid prior models for image denoising. Extensive experiments on three real-world noisy image datasets, including a new dataset constructed by us by different types of cameras and camera settings, demonstrated that our proposed method achieves much better performance than state-of-the-art image denoising methods in terms of both quantitative measure and visual perceptual quality.

\appendices
\section{Closed-Form Solution of the Weighted Sparse Coding Problem (\ref{equ7})}

For notation simplicity, we ignore the indices $n,m,t$ in problem (\ref{equ7}).\ It turns into the following weighted sparse coding problem:
\vspace{-5mm}
\begin{equation}\label{equ14}
\vspace{-2mm}
\min\nolimits_{\bm{\alpha}}\|\mathbf{y}-\bm{D}\bm{\alpha}\|_{2}^{2}+\sum\nolimits_{j=1}^{3p^2}\lambda_{j}|\bm{\alpha}_{j}|.
\end{equation}
Since $\bm{D}$ is an orthogonal matrix, problem (\ref{equ14}) is equivalent to:
\vspace{-2mm}
\begin{equation}\label{equ15}
\vspace{-2mm}
\min\nolimits_{\bm{\alpha}}\|\bm{D}^{T}\mathbf{y}-\bm{\alpha}\|_{2}^{2}+\sum\nolimits_{j=1}^{3p^2}\lambda_{j}|\bm{\alpha}_{j}|.
\end{equation}
For simplicity, we denote $\mathbf{z} = \mathbf{D^{T}y}$.\ Here we have $\lambda_{j}>0$, $j=1,...,3p^2$, then problem (\ref{equ15}) can be written as:
\vspace{-2mm}
\begin{equation}\label{equ16}
\vspace{-2mm}
\min\nolimits_{\boldsymbol{\alpha}}\sum\nolimits_{j=1}^{3p^2}((\mathbf{z}_{j}-\bm{\alpha}_{j})^{2}+\lambda_{j}|\bm{\alpha}_{j}|).
\end{equation}
The problem (\ref{equ16}) is separable w.r.t. each $\bm{\alpha}_{j}$ and hence can be simplified to $3p^2$ independent scalar minimization problems:
\vspace{-2mm}
\begin{equation}\label{equ17}
\vspace{-2mm}
\min\nolimits_{\bm{\alpha}_{j}}(\mathbf{z}_{j}-\bm{\alpha}_{j})^{2}+\lambda_{j}|\bm{\alpha}_{j}|,
\end{equation}
where $j=1,...,3p^2$. Taking derivative of $\boldsymbol{\alpha}_{j}$ in problem (\ref{equ17}) and setting the derivative to be zero. There are two cases for the solution.

(a) If $\bm{\alpha}_{j}\ge 0$, we have 
$
2(\bm{\alpha}_{j}-\mathbf{z}_{j})+\lambda_{j}=0,
$ and the solution is
$
\hat{\bm{\alpha}}_{j}=\mathbf{z}_{j}-\frac{\lambda_{j}}{2} \ge 0.
$
So $\mathbf{z}_{j}\ge\frac{\lambda_{j}}{2}> 0$, and the solution $\hat{\bm{\alpha}}_{j}$ can be written as
$
\hat{\bm{\alpha}}_{j}=\text{sgn}(\mathbf{z}_{j})*(|\mathbf{z}_{j}|-\frac{\lambda_{j}}{2}),
$
where $\text{sgn}(\bullet)$ is the sign function. 

(b) If $\bm{\alpha}_{j}< 0$, we have
$
2(\bm{\alpha}_{j}-\mathbf{z}_{j})-\lambda_{j}=0
$
and the solution is
$
\hat{\bm{\alpha}}_{j}=\mathbf{z}_{j}+\frac{\lambda_{j}}{2} < 0.
$
So $\mathbf{z}_{j}<-\frac{\lambda_{j}}{2}< 0$, and the solution $\hat{\bm{\alpha}}_{j}$ can be written as
$
\hat{\bm{\alpha}}_{j}=\text{sgn}(\mathbf{z}_{j})*(-\mathbf{z}_{j}-\frac{\lambda_{j}}{2})=\text{sgn}(\mathbf{z}_{j})*(|\mathbf{z}_{j}|-\frac{\lambda_{j}}{2}).
$

In summary, we have the final solution of the weighted sparse coding problem (\ref{equ14}) as:
\vspace{-1mm}
\begin{equation}\label{equ18}\vspace{-1mm}
\hat{\bm{\alpha}}= \text{sgn}(\mathbf{D^{T}y})\odot \text{max}(|\mathbf{D^{T}y}|-\bm{\lambda},0),
\end{equation}
where $\bm{\lambda} = \frac{1}{2}[\lambda_{1},\lambda_{2},...,\lambda_{3p^2}]^{\top}$ is the vector of regularization parameter and $\odot$ means element-wise multiplication.

\section{Proof of the Theorem \ref{th1}}

Let $\mathcal{A}\in \mathbb{R}^{(3p^2-r)\times M},\mathcal{Y}\in \mathbb{R}^{3p^2\times M}$ be two given data matrices. Denote by $\mathcal{E}\in\mathbb{R}^{3p^2\times r}$ the external subdictionary and $\mathcal{D}\in\mathbb{R}^{3p^2\times (3p^2-r)}$ the internal subdictionary. For simplicity, we assume $3p^2\ge M$. The problem in \textbf{Theorem \ref{th1}} is as follows:
\begin{equation}\label{equ19}
\begin{split}
\hat{\mathcal{D}}
=
&
\arg\min\nolimits_{\mathcal{D}}\|\mathcal{Y}-\mathcal{D}\mathcal{A}\|_{F}^{2}
\\
&
\text{s.t.}
\quad
\mathcal{D}^{\top}\mathcal{D} = \bm{I}_{(3p^2-r)\times (3p^2-r)}, \mathcal{E}^{\top}\mathcal{D} = \bm{0}_{r\times (3p^2-r)}.
\end{split}
\end{equation} 

\begin{proof}
We firstly prove the necessary condition.
Since $\mathcal{D}^{\top}\mathcal{D} = \bm{I}_{(3p^2-r)\times (3p^2-r)}$, we have
\begin{equation}\label{equ20}
\begin{split}
\hat{\mathcal{D}}
&
=
\arg\min\nolimits_{\mathcal{D}}\|\mathcal{Y}-\mathcal{D}\mathcal{A}\|_{F}^{2}
=
\arg\max\nolimits_{\mathcal{D}}\text{Tr}(\mathcal{A}\mathcal{Y}^{\top}\mathcal{D})
\\
&
\text{s.t.}
\quad
\mathcal{D}^{\top}\mathcal{D} = \bm{I}_{(3p^2-r)\times (3p^2-r)}, \mathcal{E}^{\top}\mathcal{D} = \bm{0}_{r\times (3p^2-r)}.
\end{split}
\end{equation}
The Lagrange function is
$
\mathcal{L}
=
\text{Tr}(\mathcal{A}\mathcal{Y}^{\top}\mathcal{D})
-
\text{Tr}(\Gamma_{1}(\mathcal{D}^{\top}\mathcal{D} - \bm{I}_{(3p^2-r)\times (3p^2-r)}))
-
\text{Tr}(\Gamma_{2}(\mathcal{D}^{\top}\mathcal{E}))
$,
where $\Gamma_{1}$ and $\Gamma_{2}$ are the Lagrange multipliers. Take the derivative of $\mathcal{L}$ w.r.t. $\mathcal{D}$ and set it to be matrix $\bm{0}$ of conformal dimensions, we can get
\begin{equation}\label{equ21}
\partial\mathcal{L}/\partial\mathcal{D} 
=
\mathcal{Y}\mathcal{A}^{\top}
-
\mathcal{D}(\Gamma_{1}+\Gamma_{1}^{\top})
-
\mathcal{E}\Gamma_{2}^{\top}
=
\bm{0}_{3p^2\times (3p^2-r)}.
\end{equation}
Since $\mathcal{D}^{\top}\mathcal{D}=\bm{I}_{(3p^2-r)\times (3p^2-r)}$ and $\mathcal{E}^{\top}\mathcal{D} = \bm{0}_{3p^2\times (3p^2-r)}$, by left multiplying both sides of the Eq. (\ref{equ22}) by $\mathcal{E}^{\top}$, we have 
\begin{equation}\label{equ22}
\mathcal{E}^{\top}\mathcal{Y}\mathcal{A}^{\top}
=
\Gamma_{2}^{\top}.
\end{equation}
Put the Eq. (\ref{equ22}) back into Eq. (\ref{equ21}), we have 
\begin{equation}\label{equ23}
(\bm{I}_{3p^2\times 3p^2}-\mathcal{E}\mathcal{E}^{\top})\mathcal{Y}\mathcal{A}^{\top}
=
\mathcal{D}(\Gamma_{1}+\Gamma_{1}^{\top}).
\end{equation}
Right multiplying both sides of Eq. (\ref{equ23}) by $\mathcal{D}^{\top}$, we have
\begin{equation}\label{equ24}
(\bm{I}_{3p^2\times 3p^2}-\mathcal{E}\mathcal{E}^{\top})\mathcal{Y}\mathcal{A}^{\top}\mathcal{D}^{\top}
=
\mathcal{D}(\Gamma_{1}+\Gamma_{1}^{\top})\mathcal{D}^{\top}
.
\end{equation}
This shows that $(\bm{I}_{3p^2\times 3p^2}-\mathcal{E}\mathcal{E}^{\top})\mathcal{Y}\mathcal{A}^{\top}\mathcal{D}^{\top}$ is a symmetric matrix of order $3p^2\times 3p^2$. Then we perform economy (or reduced) singular value decomposition (SVD)  \cite{eckart1936approximation} on $(\bm{I}_{3p^2\times 3p^2}-\mathcal{E}\mathcal{E}^{\top})\mathcal{Y}\mathcal{A}^{\top}=\mathcal{U}\Sigma\mathcal{V}^{\top}$,
there is
\begin{equation}\label{equ25}
(\bm{I}_{3p^2\times 3p^2}-\mathcal{E}\mathcal{E}^{\top})\mathcal{Y}\mathcal{A}^{\top}\mathcal{D}^{\top}
=
\mathcal{U}\Sigma\mathcal{V}^{\top}\mathcal{D}^{\top}
=
\mathcal{D}\mathcal{V}\Sigma\mathcal{U}^{\top}.
\end{equation}
Hence, we have $\mathcal{U}=\mathcal{D}\mathcal{V}$, or equivalently $\hat{\mathcal{D}}=\mathcal{U}\mathcal{V}^{\top}$. The necessary condition is proved. 

Now we prove the sufficient condition. If $\hat{\mathcal{D}}=\mathcal{U}\mathcal{V}^{\top}$, then $\hat{\mathcal{D}}^{\top}\hat{\mathcal{D}}=\bm{I}_{(3p^2-r)\times (3p^2-r)}$. To prove $\mathcal{E}^{\top}\hat{\mathcal{D}}=\bm{0}_{3p^2\times (3p^2-r)}$, we left multiply both sides of Eq. (\ref{equ25}) by $\mathcal{E}^{\top}$ and have  
$
\bm{0}_{3p^2\times (3p^2-r)}
=
\mathcal{E}^{\top}(\bm{I}_{3p^2\times 3p^2}-\mathcal{E}\mathcal{E}^{\top})\mathcal{Y}\mathcal{A}^{\top}\hat{\mathcal{D}}^{\top}
=
\mathcal{E}^{\top}\mathcal{U}\Sigma\mathcal{V}^{\top}\hat{\mathcal{D}}^{\top}
=
\mathcal{E}^{\top}\mathcal{U}\Sigma\mathcal{U}^{\top}
$
.
It means that $\mathcal{E}^{\top}\mathcal{U}\Sigma\mathcal{U}^{\top}=\bm{0}_{3p^2\times 3p^2}$. This only happens when $\mathcal{E}^{\top}\mathcal{U}=\bm{0}_{3p^2\times (3p^2-r)}$ since $\text{rank}(\Sigma)=3p^2-r$ and $\mathcal{U}\Sigma\mathcal{U}^{\top}$ is positive definite. Then $\mathcal{E}^{\top}\hat{\mathcal{D}}=\mathcal{E}^{\top}\mathcal{U}\mathcal{V}^{\top}=\bm{0}_{3p^2\times (3p^2-r)}$. 

Finally we prove that $\hat{\mathcal{D}}=\mathcal{U}\mathcal{V}^{\top}$ is the solution of
\begin{equation}\label{equ26}
\begin{split}
\hat{\mathcal{D}}
=
\arg\min\nolimits_{\mathcal{D}}
\|\mathcal{Y}-\mathcal{D}\mathcal{A}\|_{F}^{2}
=
\arg\max\nolimits_{\mathcal{D}}
\text{Tr}(\mathcal{Y}^{\top}\mathcal{D}\mathcal{A}).
\end{split}
\end{equation}
Note that by cyclic perturbation which retains the trace unchanged and due to $\mathcal{E}^{\top}\hat{\mathcal{D}}=\bm{0}_{3p^2\times (3p^2-r)}$, we have 
$
\text{Tr}(\mathcal{Y}^{\top}\hat{\mathcal{D}}\mathcal{A})
=
\text{Tr}(\mathcal{Y}\mathcal{A}^{\top}\hat{\mathcal{D}}^{\top})
=
\text{Tr}((\bm{I}_{3p^2\times 3p^2}-\mathcal{E}\mathcal{E}^{\top})\mathcal{Y}\mathcal{A}^{\top}\hat{\mathcal{D}}^{\top})
=
\text{Tr}(\mathcal{U}\Sigma\mathcal{V}^{\top}\mathcal{V}\mathcal{U}^{\top})
=
\text{Tr}(\Sigma).
$
For every $\mathcal{D}$ satisfying that $\mathcal{D}^{\top}\mathcal{D} = \bm{I}_{(3p^2-r)\times (3p^2-r)}$, $\mathcal{E}^{\top}\mathcal{D} = \bm{0}_{3p^2\times (3p^2-r)}$, we have 
$
\text{Tr}(\mathcal{Y}^{\top}\mathcal{D}\mathcal{A})
=
\text{Tr}((\bm{I}_{3p^2\times 3p^2}-\mathcal{E}\mathcal{E}^{\top})\mathcal{Y}\mathcal{A}^{\top}\mathcal{D}^{\top})
=
\text{Tr}(\mathcal{U}\Sigma\mathcal{V}^{\top}\mathcal{D}^{\top})
=
\text{Tr}(\Sigma\mathcal{V}^{\top}\mathcal{D}^{\top}\mathcal{U})
$.
By using a generalization version \cite{TenBerge1983} of the Kristof's Theorem \cite{Kristof1970515}, we have $\text{Tr}(\mathcal{Y}^{\top}\mathcal{D}\mathcal{A})
=
\text{Tr}(\Sigma\mathcal{V}^{\top}\mathcal{D}^{\top}\mathcal{U})
\le
\text{Tr}(\Sigma)
.
$
The equality is obtained at 
$\mathcal{V}^{\top}\mathcal{D}^{\top}\mathcal{U}=\bm{I}_{(3p^2-r)\times (3p^2-r)}$, i.e., $\mathcal{D}=\mathcal{U}\mathcal{V}^{\top}=\hat{\mathcal{D}}$. This completes the proof.
\end{proof}

\section{Proof of the Theorem \ref{th2}}

Before we prove the Theorem \ref{th2}, we need firstly prove the following Lemma 1.

\emph{Lemma 1}: Let $\mathcal{E}\in\mathbb{R}^{3p^2\times r}$ be an orthogonal matrix with $\mathcal{E}^{\top}\mathcal{E}=\bm{I}_{r\times r}$, then $\text{rank}(\bm{I}_{3p^2\times 3p^2}-\mathcal{E}\mathcal{E}^{\top})\ge 3p^2-r$.

\begin{proof} Since $\text{rank}(\mathcal{E}\mathcal{E}^{\top})\le\min\{\text{rank}(\mathcal{E}),\text{rank}(\mathcal{E}^{\top})\}=r$ and $\text{rank}(\mathcal{E}\mathcal{E}^{\top})\ge\text{rank}(\mathcal{E})+\text{rank}(\mathcal{E}^{\top})-r=2r-r=r$ by Sylvester's inequality, we have $\text{rank}(\mathcal{E}\mathcal{E}^{\top})=r$. Then, $\text{rank}(\bm{I}_{3p^2\times 3p^2}-\mathcal{E}\mathcal{E}^{\top})\ge\text{rank}(\bm{I}_{3p^2\times 3p^2})-\text{rank}(\mathcal{E}\mathcal{E}^{\top})\ge 3p^2-r$. 
\end{proof}

The $\text{rank}(\Sigma)$ ($\Sigma$ is defined in Theorem \ref{th1}) depends on $\text{rank}(\bm{I}_{3p^2\times 3p^2}-\mathcal{E}\mathcal{E}^{\top})$, $\text{rank}(\mathcal{Y})$ and $\text{rank}(\mathcal{A})$. Note that $\text{rank}(\mathcal{Y})\ge M$ and $\text{rank}(\mathcal{A})\ge \min\{3p^2,M\}$ and $\text{rank}(\bm{I}_{3p^2\times 3p^2}-\mathcal{E}\mathcal{E}^{\top})\ge 3p^2-r$. Hence, $\text{rank}(\Sigma)\le\min\{3p^2-r,M\}$.

Now we prove the Theorem \ref{th2}:
\begin{proof} 
a) If $(\bm{I}_{3p^2\times 3p^2}-\mathcal{E}\mathcal{E}^{\top})\mathcal{Y}\mathcal{A}^{\top}\in\mathbb{R}^{3p^2\times (3p^2-r)}$ is nonsingular, i.e., $\text{rank}(\Sigma)=3p^2-r$, $\Sigma$ may have distinct or multiple non-zero singular values. In the SVD \cite{eckart1936approximation} of $(\bm{I}_{3p^2\times 3p^2}-\mathcal{E}\mathcal{E}^{\top})\mathcal{Y}\mathcal{A}^{\top}
=
\mathcal{U}\Sigma\mathcal{V}^{\top}$, the singular vectors in $\mathcal{U}$ and $\mathcal{V}$
can be determined up to orientation. Hence, we can reformulate it as 
\begin{equation}\label{equ27}
(\bm{I}_{3p^2\times 3p^2}-\mathcal{E}\mathcal{E}^{\top})\mathcal{Y}\mathcal{A}^{\top}
=
\mathcal{U}^{*}\mathcal{K}_{u}\Sigma\mathcal{K}_{v}(\mathcal{V}^{*})^{\top},
\end{equation}
where $\mathcal{U}^{*}\in \mathbb{R}^{3p^2\times (3p^2-r)}$ and $\mathcal{V}^{*}\in \mathbb{R}^{(3p^2-r)\times (3p^2-r)}$ are arbitrarily orientated singular vectors of $\mathcal{U}$ and $\mathcal{V}$, respectively. The $\mathcal{K}_{u}$ and $\mathcal{K}_{v}$ are diagonal matrices with $+1$ or $-1$ as diagonal elements in arbitrary distribution. $\Sigma\in \mathbb{R}^{(3p^2-r)\times (3p^2-r)}$ is a diagonal matrix with singular values in non-increasing order, i.e., $\Sigma_{11}\ge\Sigma_{22}\ge...\ge\Sigma_{(3p^2-r)(3p^2-r)}\ge0$. If we fix $\mathcal{K}_{u}$, then $\mathcal{K}_{v}$ is uniquely determined to meet the above requirements of $\Sigma$. If the orientations of the singular vectors of $\mathcal{U}^{*}$ are fixed, then $\mathcal{U}=\mathcal{U}^{*}\mathcal{K}_{u}$ is determined, so do the orientations of the singular vectors of $\mathcal{V}^{*}$ and $\mathcal{V}^{\top}=\mathcal{K}_{v}(\mathcal{V}^{*})^{\top}$. In this case, the solution of $\hat{\mathcal{D}}=\mathcal{U}\mathcal{V}^{\top}=\mathcal{U}^{*}\mathcal{K}_{u}\mathcal{K}_{v}(\mathcal{V}^{*})^{\top}$ is unique. When $\Sigma$ has multiple singular values, the unique solution of $\hat{\mathcal{D}}$ can be proved in a similar way. 

b) If $(\bm{I}_{3p^2\times 3p^2}-\mathcal{E}\mathcal{E}^{\top})\mathcal{Y}\mathcal{A}^{\top}$ is singular, i.e.,  $0\le \text{rank}(\Sigma)< 3p^2-r$, and $\Sigma$ has $3p^2-r-\text{rank}(\Sigma)$ (at least one) zero singular values. The discussion in a) can still be applied to the singular vectors corresponding to the nonzero singular values, and the production of these singular vectors in $\mathcal{U}$ and $\mathcal{V}$ is still unique. However, the singular vectors corresponding to the zero singular values could be in arbitrary orientations as long as they satisfy the conditions of $\mathcal{U}^{\top}\mathcal{U}=\mathcal{V}^{\top}\mathcal{V}=\mathcal{V}\mathcal{V}^{\top}=\bm{I}_{(3p^2-r)\times (3p^2-r)}$. Since $\mathcal{U}\in \mathbb{R}^{3p^2\times (3p^2-r)}$, $\mathcal{U}\mathcal{U}^{\top}$ no longer equals to the identity matrix of order $3p^2\times 3p^2$. From Eq. (\ref{equ25}), we have
\begin{equation}\label{equ28}
\mathcal{U}\Sigma\mathcal{V}^{\top}\mathcal{D}^{\top}
=
\mathcal{D}\mathcal{V}\Sigma\mathcal{U}^{\top}
\vspace{-2mm}
\end{equation}
Right multiplying both sides of Eq.\ (\ref{equ28}) by $\mathcal{D}\mathcal{V}$ and left multiplying each side by $\mathcal{U}^{\top}$, we have
\vspace{-2mm}
\begin{equation}\label{equ29}
\Sigma
=
\mathcal{U}^{\top}\mathcal{D}\mathcal{V}\Sigma\mathcal{U}^{\top}\mathcal{D}\mathcal{V}
\vspace{-2mm}
\end{equation}
Hence, $\Delta=\mathcal{U}^{\top}\mathcal{D}\mathcal{V}\in\mathbb{R}^{(3p^2-r)\times (3p^2-r)}$ is a diagonal matrix, the diagonal elements of which are 
\vspace{-1mm}
\begin{displaymath}
\Delta_{ii}= \left\{ \begin{array}{ll}
1 & \textrm{if $1\le i\le \text{rank}(\Sigma)$};\\
\pm 1 & \textrm{if $\text{rank}(\Sigma)< i \le 3p^2-r$}.\\
\end{array} \right.
\vspace{-1mm}
\end{displaymath}
Thus, we have $\mathcal{D}=\mathcal{U}\Delta\mathcal{V}^{\top}$. That is, if $\text{rank}(\Sigma)<3p^2-r$, once we get the solution of $\hat{\mathcal{D}}=\mathcal{U}\mathcal{V}^{\top}$ in problem (\ref{equ19}), $\mathcal{D}=\mathcal{U}\Delta\mathcal{V}^{\top}$ with suitable $\Delta$ is also the solution of problem (\ref{equ19}). In fact, the number of solutions $\hat{\mathcal{D}}$ for problem (\ref{equ19}) is $2^{3p^2-r-\text{rank}(\Sigma)}$ given fixed $\mathcal{U}$ and $\mathcal{V}$.
\end{proof}

{
\small
\bibliographystyle{unsrt}
\bibliography{egbib}
}

\vspace{-10mm}
\begin{IEEEbiography}[{\includegraphics[width=1in,height=1.25in,clip,keepaspectratio]{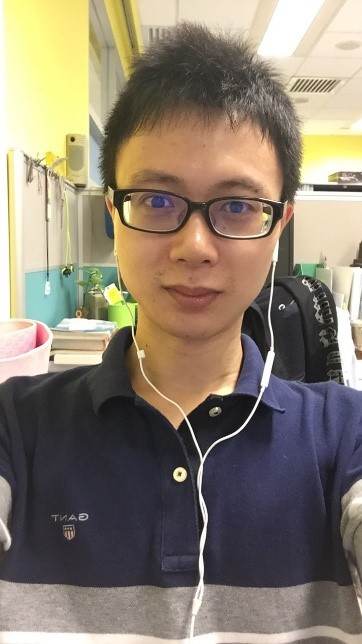}}]{Jun Xu}
 received the B.Sc. degree in Pure Mathematics and the M. Sc. Degree in Information and Probability both from the School of Mathematics Science, Nankai University, China, in 2011 and 2014, respectively. He is currently pursuing the Ph.D. degree in the Department of Computing, The Hong Kong Polytechnic University. His research interests include image restoration, subspace clustering, sparse and low rank models.
\end{IEEEbiography}

\vspace{-10mm}
\begin{IEEEbiography}[{\includegraphics[width=1in,height=1.25in,clip,keepaspectratio]{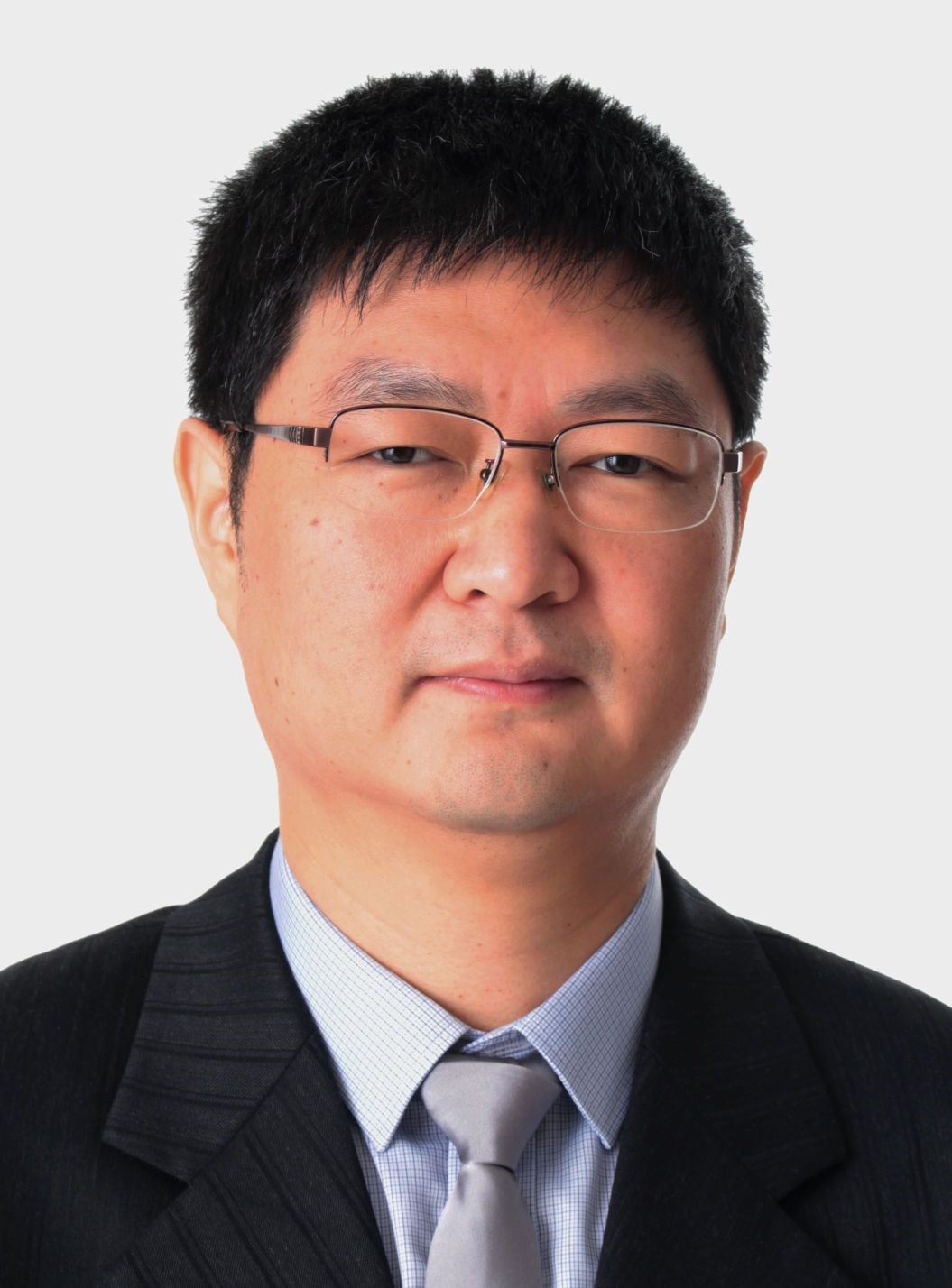}}]{Lei Zhang} (M'04, SM'14, F'18) received his B.Sc. degree in 1995 from Shenyang Institute of Aeronautical Engineering, Shenyang, P.R. China, and M.Sc. and Ph.D degrees in Control Theory and Engineering from Northwestern Polytechnical University, Xi'an, P.R. China, respectively in 1998 and 2001, respectively. From 2001 to 2002, he was a research associate in the Department of Computing, The Hong Kong Polytechnic University. From January 2003 to January 2006 he worked as a Postdoctoral Fellow in the Department of Electrical and Computer Engineering, McMaster University, Canada. In 2006, he joined the Department of Computing, The Hong Kong Polytechnic University, as an Assistant Professor. Since July 2017, he has been a Chair Professor in the same department. His research interests include Computer Vision, Pattern Recognition, Image and Video Analysis, and Biometrics, etc. Prof. Zhang has published more than 200 papers in those areas. As of 2018, his publications have been cited more than 30,000 times in the literature. Prof. Zhang is an Associate Editor of \textsl{IEEE Trans. on Image Processing}, \textsl{SIAM Journal of Imaging Sciences} and \textsl{Image and Vision Computing}, etc. He is a ``Clarivate Analytics Highly Cited Researcher'' from 2015 to 2017. More information can be found in his homepage \url{http://www4.comp.polyu.edu.hk/~cslzhang/}.
\end{IEEEbiography}

\vspace{-10mm}
\begin{IEEEbiography}[{\includegraphics[width=1in,height=1.25in,clip,keepaspectratio]{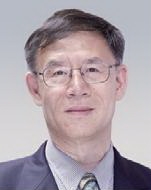}}]{David Zhang} (F'08) received the degree in computer science from Peking University, the M.Sc. degree in 1982, and the Ph.D. degree in computer science from the Harbin Institute of Technology (HIT), in 1985, respectively. From 1986 to 1988, he was a Post-Doctoral Fellow with Tsinghua University and an Associate Professor with the Academia Sinica, Beijing. In 1994, he received the second Ph.D. degree in electrical and computer engineering from the University of Waterloo, Ontario, Canada. He is currently the Chair Professor with the Hong Kong Polytechnic University, since 2005, where he is the Founding Director of the Biometrics Research Centre (UGC/CRC) supported by the Hong Kong SAR Government in 1998. He is a Croucher Senior Research Fellow, Distinguished Speaker of the IEEE Computer Society, and a Fellow of IAPR. So far, he has published over 20 monographs, over 400 international journal papers and over 40 patents from USA/Japan/HK/China. He was selected as a Highly Cited Researcher in Engineering by Thomson Reuters in 2014, 2015, and 2016, respectively. He also serves as a Visiting Chair Professor with Tsinghua University and an Adjunct Professor with Peking University, Shanghai Jiao Tong University, HIT, and the University of Waterloo. He is Founder and Editor-in-Chief, International Journal of Image and Graphics, the Founder and the Series Editor, Springer International Series on Biometrics (KISB); Organizer, International Conference on Biometrics Authentication, an Associate Editor for over ten international journals including the IEEE Transactions and so on.
\end{IEEEbiography}

\end{document}